\documentclass[10pt,journal,compsoc]{IEEEtran}
\usepackage{tikz,xcolor,hyperref}
\usepackage{comment}
\usepackage{ragged2e}
\usepackage{amsopn}
\usepackage{amsmath}
\usepackage{amssymb}
\usepackage{multirow}
\usepackage{algorithm}
\usepackage{algorithmic}
\usepackage{colortbl}
\usepackage{cite}
\usepackage{booktabs}
\usepackage{subfigure}
\definecolor{lime}{HTML}{A6CE39}
\DeclareRobustCommand{\orcidicon}{%
	\begin{tikzpicture}
	\draw[lime, fill=lime] (0,0)
	circle [radius=0.16]
	node[white] {{\fontfamily{qag}\selectfont \tiny ID}};	\draw[white, fill=white] (-0.0625,0.095)
	circle [radius=0.007];	\end{tikzpicture}
	\hspace{-2mm}}
\foreach \x in {A, ..., Z}{%
	\expandafter\xdef\csname orcid\x\endcsname{\noexpand\href{https://orcid.org/\csname orcidauthor\x\endcsname}{\noexpand\orcidicon}}
	}

\hypersetup{
colorlinks=true,
linkcolor=black
}

\begin{document}

\title{PKSS-Align: Robust Point Cloud Registration on Pre-Kendall Shape Space}

\author{Chenlei~Lv\orcidA{},~\IEEEmembership{Member,~IEEE,} 
        Hui~Huang$^*$\orcidB{},~\IEEEmembership{Senior Member,~IEEE,}
\IEEEcompsocitemizethanks{

\IEEEcompsocthanksitem Chenlei~Lv and Hui~Huang are with Shenzhen University, College of Computer Science and Software Engineering, Visual Computing Research Center, Shenzhen 518060, China. The corresponding author is Hui~Huang (hhzhiyan@gmail.com).}
\thanks{Manuscript submitted July, 2025}}

\markboth{}%
{Lv \MakeLowercase{\textit{et al.}}: PKSS-Align: Robust Point Cloud Registration on Pre-Kendall Shape Space}

\IEEEtitleabstractindextext{%
\begin{abstract}

Point cloud registration is a classical topic in the field of 3D Vision and Computer Graphics. Generally, the implementation of registration is typically sensitive to similarity transformations (translation, scaling, and rotation), noisy points, and incomplete geometric structures. Especially, the non-uniform scales and defective parts of point clouds increase probability of struck local optima in registration task. In this paper, we propose a robust point cloud registration PKSS-Align that can handle various influences, including similarity transformations, non-uniform densities, random noisy points, and defective parts. The proposed method measures shape feature-based similarity between point clouds on the Pre-Kendall shape space (PKSS), \textcolor{black}{which is a shape measurement-based scheme and doesn't require point-to-point or point-to-plane metric.} The employed measurement can be regarded as the manifold metric that is robust to various representations in the Euclidean coordinate system. Benefited from the measurement, the transformation matrix can be directly generated for point clouds with mentioned influences at the same time. The proposed method does not require data training and complex feature encoding. Based on a simple parallel acceleration, it can achieve significant improvement for efficiency and feasibility in practice. Experiments demonstrate that our method outperforms the relevant state-of-the-art methods. Project link: \href{https://github.com/vvvwo/PKSS-Align}{\color{blue}https://github.com/vvvwo/PKSS-Align}.

\end{abstract}

}

\maketitle

\IEEEdisplaynontitleabstractindextext

\IEEEpeerreviewmaketitle

\IEEEraisesectionheading{\section{Introduction}\label{sec:introduction}}

\begin{figure}
  \centering
  \includegraphics[width=\linewidth]{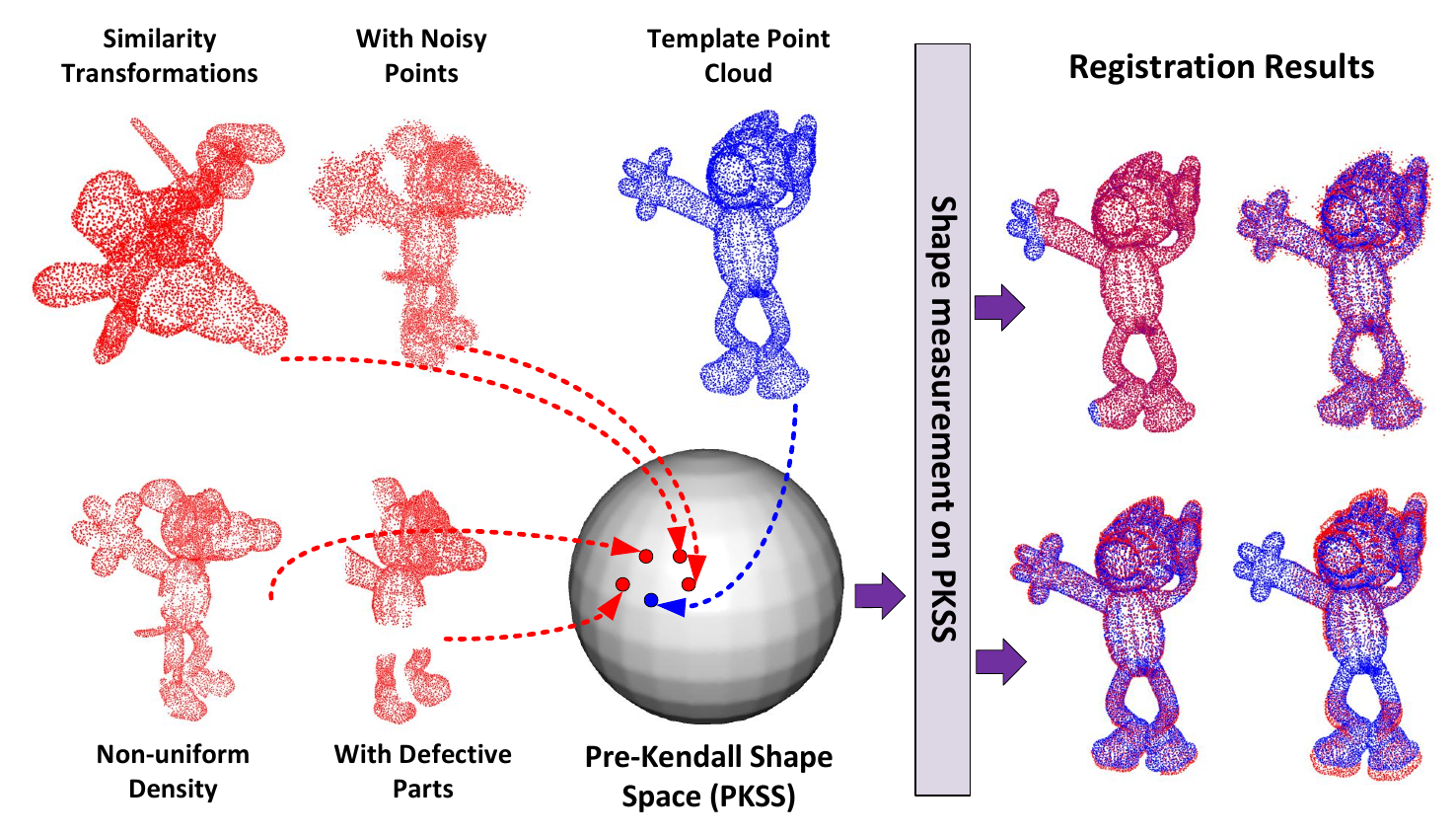}
  \vspace{-4mm}
  \caption{Instances of point cloud registration based on Pre-Kendall shape space (PKSS). The source point clouds with different similarity transformations, non-uniform densities, noisy points, and defective parts are mapped onto the PKSS. Based on the shape measurement on PKSS, such point clouds are aligned to the template ones.}
   \vspace{-4mm}
  \label{f1}
\end{figure}

\IEEEPARstart{A}{s} an important topic in 3D vision, point cloud registration has been studied for many years. The target of the registration is to align point clouds with accurate point-based correspondence and semantic consistency. It is useful for many applications such as simultaneous localization and mapping (SLAM), autopilot system, 3D reconstruction, etc. To implement the registration, some challenges should be considered, including influences of similarity transformations, non-uniform densities, random noisy points, and incomplete geometric structures. Such influences are produced by complex external conditions during the point cloud scanning. Some instances are shown in Fig.~\ref{f1}.

To handle the challenges, there are several main technical routes, including distance-based registration, geometric feature-based alignment, and deep encoding-based correspondence. The distance-based registration is to match point clouds based on point-to-point~\cite{besl1992method} or point-to-plane~\cite{park2003accurate} distance. Following the descriptions of Wahba problem~\cite{wahba1965least}, once the correspondence between two vectors is established, an optimal transformation can be obtained by singular value decomposition, which can be regarded as the transformation matrix for registration. It is convenient to be implemented and does not require complex feature analysis. However, the registration depends on initial poses of point clouds that are typically unordered. It significantly reduces the function of singular value decomposition for vector alignment. The drawback increases the probability of local optima occurring, especially for the point clouds with non-uniform scales and defective parts. The geometric feature-based alignment~\cite{rusu2009fast} can be used to improve the accuracy. The transformation matrix is estimated directly and independent of the initial pose. It also can be concluded as the correspondence-based registration that is consistent to the basic framework of Horn's method~\cite{horn1988closed}. Intuitively, the performance of the alignment is sensitive to the quality of selected geometric feature. In addition, the feature extraction reduces the efficiency in practice. Following the development of deep learning, some researchers propose related frameworks~\cite{aoki2019pointnetlk}\cite{sarode2019pcrnet} to implement point cloud registration. Such frameworks encode point clouds into learnable deep-features for efficient calculation. However, the frameworks are sensitive to the training dataset with various similarity transformations that increase the probability of the local optimum.

Considering the disadvantages of the mentioned technical routes, a robust registration solution should satisfy the characteristics: \textbf{accurate and robust measurement} for registration status, \textcolor{black}{\textbf{independent to the feature-based or strict point-based correspondence} (for pose alignment, it does not rely on precise geometric features or points, and robust to low-quality data such as noise and random distributions.)}, and \textbf{global searching} to avoid the local optimum. In our previous work~\cite{lv2023kss}, we have proven that the Kendall shape space theory~\cite{kendall1984shape} can support an effective solution for registration \textcolor{black}{that is a shape measurement-based scheme}. However, the solution leaves two limitations: lower computational efficiency and restricted registration for point clouds with different scales and defective parts. For shape measurement, the employed Hausdorff metric cannot be considered a manifold metric, which reduce the robustness in registration task.

\begin{figure*}
  \centering
  \includegraphics[width=\linewidth]{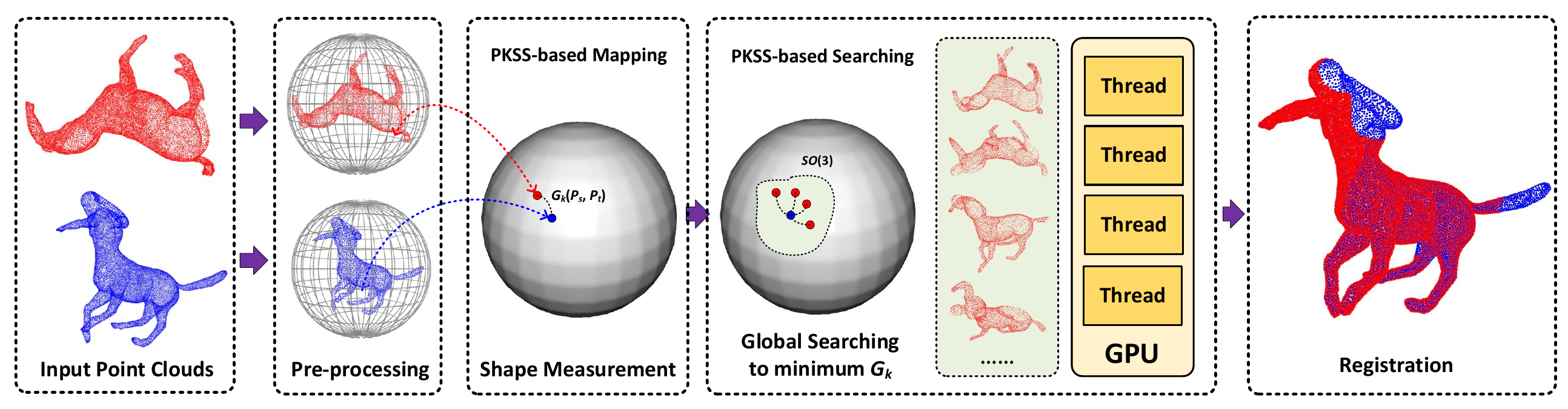}
  \vspace{-4mm}
  \caption{The pipeline of PKSS-Align. Firstly, the pre-processing is used to normalize point clouds and remove outliers; secondly, the shape measurement estimates the pose similarity between point clouds in current state; finally, the global searching is to find the best transfer matrix based on the shape measurements from the candidate transfer set in parallel.}
  \vspace{-4mm}
  \label{f2}
\end{figure*}

In this paper, we propose a new method PKSS-Align that is an enhanced version of KSS-ICP \cite{lv2023kss}. It is implemented on Pre-Kendall shape space (PKSS) with main components: PKSS-based mapping, PKSS-based shape measurement, and global searching scheme in $SO(3)$. Firstly, the PKSS-based mapping is implemented to reduce influences of different translations and scales. Next, the PKSS-based shape measurement is proposed to provide a manifold metric for alignment. The measurement inherits the advantages of robustness to similarity transformations while addressing the limitations of KSS-ICP. Finally, the global searching scheme in $SO(3)$ with parallel acceleration is employed to generate registration result. It can efficiently handle the different scales and defective parts at the same time. Compared to the previous version, the new method has significantly improved in terms of robustness and computational efficiency. The pipeline is shown in Fig.~\ref{f2} and the contributions are summarized as:

\begin{itemize}
    \item We provide a PKSS-based mapping to represent point cloud on PKSS. The influences of scaling and translation are initially reduced that is helpful for following alignment.
    \item We present a PKSS-based shape measurement to describe the align status between point clouds. The measurement does not depend on point-based metric and complex geometric features. It provides more accurate and robust quantitative analysis for various poses of point clouds. Specifically, it overcomes the limitations of Hausdorff metric in previous version.
    \item We design a global searching scheme in $SO(3)$ with parallel acceleration. The scheme searches larger solution space to avoid potential local optimum for registration. In addition, it provides a practical solution for point clouds with defective parts and different scales at the same time.
\end{itemize}

The rest of the paper is organized as follows. In Sec. 2, we review existing classical methods for point cloud registration. In Sec. 3, we introduce the details of PKSS-based mapping, followed by the implementations of PKSS-based shape measurement and global searching scheme in Sec. 4 and Sec. 5. We demonstrate the effectiveness and efficiency of our method with extensive experimental evidence in Sec. 6, and Sec. 7 concludes the paper.

\section{Related Works}

The registration task concludes a large number of solutions. One classification divides the solutions into two categories: correspondence-based and correspondence-free. Based on our study, we have found that many solutions incorporate a combination of methods to implement registration. Therefore, we employ a new classification to facilitate comprehension, including distance metric-based, feature-based, and deep encoding-based registrations.

\textbf{Distance metric-based registration} methods achieve registration results by point-to-point or point-to-plane distance optimization. The Iterative Closest Point (ICP)~\cite{besl1992method} and its variants~\cite{fitzgibbon2003robust, granger2002multi, ying2009scale, jian2005robust, yang2019point, Zhang2022FastICP, lv2023kss} belong to this category. More related works are discussed in reviews~\cite{makela2002review}\cite{rusinkiewicz2001efficient}. In Sec.~\ref{sec:introduction}, we have introduced that such methods construct local searching strategy that depends on the initial poses of point clouds. The drawback is that the registration traps into the local optimum with high probability. To avoid the local optimum, some methods utilize the Branch-and-Bound (BnB) scheme~\cite{olsson2008branch} to implement global searching in $SO(3)$, including $L2$ error optimization~\cite{li20073d}, stereographic projection~\cite{parra2014fast}, consensus set maximization~\cite{bazin2012globally}, camera pose alignment~\cite{enqvist2008robust}, and globally optimal solution(Go-ICP)~\cite{yang2013go}. However, most of them are sensitive to point clouds with non-uniform scales and defective parts. Even the BnB scheme provides the global searching implementation, the accuracy of the registration also can not be guaranteed for complex similarity transformations.

\textbf{Feature-based registration} methods build the correspondence of point clouds based on geometric features or shape descriptors. Such methods estimate the transformation matrix by feature alignment directly. In theory, the feature alignment keeps better geometric consistency during the registration process. According to the related features, some classical methods establish their implementations, including Normal Distributions Transform (NDT)~\cite{biber2003normal, serafin2015nicp}, Shape Context~\cite{belongie2002shape,frome2004recognizing}, Sub-maps~\cite{guan2009registration}, Rotational Projection Statistics features~\cite{guo2014accurate}, Covariance Matrices~\cite{tabia2015covariance}, 
Point Feature Histograms (PFH)~\cite{rusu2008aligning, rusu2009fast, yang2016fast}, Second Order Spatial Compatibility Measure (SC$^2$-PCR)~\cite{chen2022sc2}, Maximal Cliques~\cite{zhang20233d}, \textcolor{black}{Inlier Confidence Calibration~\cite{yuan2024inlier}, and Progressive Distance Extension~\cite{liu2024extend}}. Generally, the main drawback of such methods is that the performance of registration is sensitive to the quality of the selected feature. Noisy points and defective parts in point clouds reduce the accuracy and robustness of the feature inevitably. For point clouds with large volumes, the huge calculation of feature extraction also affects the practicality of the methods.

\textbf{Deep encoding-based registration} is becoming more popular recently. Comparing to the traditional feature-based registration, the deep encoding-based methods encode the learnable deep-features~\cite{qi2017pointnet}\cite{qi2017pointnet++} to represent the point cloud. Benefited from the large dataset-based training and efficient computing, such methods achieve the balance between efficiency and robustness for registration. The methods include PointNetLK~\cite{aoki2019pointnetlk}, Deep ICP~\cite{lu2019deepicp}, Deep Closest Point~\cite{wang2019deep}, PRNet~\cite{wang2019prnet}, IDAM~\cite{li2019iterative}, RPM-Net~\cite{yew2020rpm}, 3DRegNet~\cite{pais20203dregnet}, DGR~\cite{choy2020deep}, PCRNet~\cite{sarode2019pcrnet}, Recurrent Closest Point~\cite{gu2022rcp}, GeoTransformer~\cite{qin2022geometric}, RoReg~\cite{wang2023roreg}, ROTBS~\cite{mei2021point}, RGM~\cite{fu2021robust}, REGTR~\cite{yew2022regtr}, \textcolor{black}{UDPReg~\cite{mei2023unsupervised}}, RoITr~\cite{yu2023rotation}, and Wednesday~\cite{jin2024multiway}. Although the methods take significant computational improvement, some defects are still exist in practice. Such frameworks learn the deep features from the  training dataset and encode the point clouds for registration. The performance is limited by the categories and distribution of the dataset. In other word, the learnable deep-features are semantically sensitive. Without reasonable pre-processing, the framework can not reduce the influence of different scales and non-uniform density of point clouds.

The fundamental problem with existing methods is that their inability to establish local feature-free measurement on unorganized point clouds with global correspondence. It becomes more challenging when there are scale differences and data incompleteness. KSS-ICP~\cite{lv2023kss} provides a feasible solution with global shape alignment property. Unfortunately, the related limitations restricts its performance in practice. The proposed PKSS-Align completely improves the shortcomings of KSS-ICP by employing new measurement and global pose searching strategy. It enables simultaneous improvement in computational speed and accuracy. In following parts, we will introduce implementation details.

\section{PKSS-based Mapping}\label{Pre-processing}

Kendall shape space~\cite{kendall1984shape} has been widely used in shape analysis. Various 3D objects can be represented by regular discrete forms and measured on the Kendall shape space. According to the Kendall theory~\cite{kendall1984shape}, the PKSS is a quotient space that removes influences of partial similarity transformations (scaling and translation). The pre-processing is used to map point clouds onto PKSS. The projection of a point cloud on PKSS can be regarded as a "Pre-shape" regular form~\cite{nava2020geodesic}. Although the reflection is still  affected by the rotation, it reduces the complexity for shape alignment that is useful for registration. Based on the property, the mapping operation is to find the regular form for point clouds by normalizing different scales and translations. Let $P$ represents an input point cloud, $P=\{x_1,...,x_n\}$, $x_i$ is a 3D point belongs to $P$. ${\boldsymbol K}_{\mathbf p\mathbf r\mathbf e}$ represents the PKSS that can be formulated as
\begin{equation}
{\boldsymbol K}_{\mathbf p\mathbf r\mathbf e}=\boldsymbol R^{m\times3}\backslash G_{st},
\end{equation}
where $G_{st}$ represent a transformation group includes scaling and translation. A point cloud-based reflection on ${\boldsymbol K}_{\mathbf p\mathbf r\mathbf e}$ has $m$ points and each one has three dimensions. In addition, the ${\boldsymbol K}_{\mathbf p\mathbf r\mathbf e}$ has removed the influences of scaling and translation with quotient group operation. Following the formulation, we provide the implementation for the mapping:
\begin{equation}
\begin{array}{c}
\boldsymbol K_{\mathbf{pre}}(P)=\{x_1-\overline x,...,x_m-\overline x\}/s(P),\\
\overline x={\textstyle\frac1m}\sum_{j=1}^mx_j,s(P)={(\sum_{j=1}^m\left\|x_j-\overline x\right\|)}^{1/2}.
\end{array}
\label{e2}
\end{equation}

In theory, a point cloud can be mapped onto the ${\boldsymbol K}_{\mathbf p\mathbf r\mathbf e}$ by Eq.~\eqref{e2}. However, there have four influencing factors that cannot be ignored, including different point numbers, non-uniform density, random outliers, and non-uniform scaling produced by defective parts. In general, the raw point clouds have different point numbers. For PKSS-based mapping, the numbers should be unified to $m$. The non-uniform density takes some unpredictable disturbances that reduces the accurate of $\overline x$. It means that the point clouds scanned from same object with different densities are not treated as the same model. To solve the problems, a point cloud resampling method~\cite{lv2021approximate} is used to uniform point numbers and densities. It can be regarded as an improved farthest point sampling with parallel acceleration. Based on the resampling, the first two influencing factors are removed. Even the point numbers are not strictly equal by resampling, our method also can achieve the reliable measurement. The details are discussed in Sec.~\ref{Shape Measurement on PKSS}.

The random outliers break the geometric structure of the point cloud. The more traditional solution for outlier culling is to compute the k-nearest neighbor (KNN). The KNN-based distance of the outlier significantly larger than ordinary points. With a specified threshold based on the distance, the outliers can be detected. However, the solution is not an adaptive method that affected by the threshold and outliers' density. We present an adaptive outlier culling method without specified threshold. The basic assumption of the method is that the ordinary points should have duality for neighbors. The outlier culling is formulated as
\begin{equation}
\{x_o\}=\{x_i\vert x_i\not\in N(x_j),\;x_j\in N(x_i)\},
\label{e3}
\end{equation}
where $\{x_o\}$ is the outlier set, $N$ represents the neighbor region. We detect the outlier set and remove it to implement the outlier culling. 

The non-uniform scaling produced by defective parts is difficult to be removed directly. Traditional scale normalization methods like PCA-based estimation~\cite{dimitrov2009bounds} may invalid for point clouds with defective parts. It cannot be solved by Eq.~\eqref{e2}. We provide the shape measurement-based solution that is discussed in the following section.

\section{PKSS-based Shape Measurement}\label{Shape Measurement on PKSS}

It has been discussed that the accurate measurement is important for registration. Benefited from the Kendall theory, the shape measurement can be computed on PKSS without complex feature learning. It reflects the shape similarity between the source point cloud and the template one. The formulation of the measurement is represented as:
\begin{equation}
G_{{\boldsymbol K}}({\boldsymbol K}_{\mathbf{pre}}(P_a), O{\boldsymbol K}_{\mathbf{pre}}(P_b))=arc\cos tr(R\Lambda),
\label{e4}
\end{equation}
\begin{equation}
{\boldsymbol K}_{\mathbf{pre}}(P_a){\boldsymbol K}_{\mathbf{pre}}(P_b)^T=U\Lambda V,\;R\in SO(3),
\label{e5}
\end{equation}
where $G_{{\boldsymbol K}}$ is the measurement, ${\boldsymbol K}_{\mathbf{pre}}(P_a)$ and ${\boldsymbol K}_{\mathbf{pre}}(P_b)$ are point clouds that mapped onto the PKSS, $O$ represents a rotation that implemented into the $P_b$. $\Lambda$ is the real diagonal matrix computed by singular value decomposition. It includes the eigenvalues of ${\boldsymbol K}_{\mathbf{pre}}(P_a){\boldsymbol K}_{\mathbf{pre}}(P_b)^T$. The measurement reflects the shape similarity between vectorized point clouds according to the eigenvalues. The registration is to find the rotation $O$ that minimizes the $G_{{\boldsymbol K}}({\boldsymbol K}_{\mathbf{pre}}(P_a),O\times {\boldsymbol K}_{\mathbf{pre}}(P_b))$. According to the Kendall theory, the minimum of the measurement is represented as
\begin{equation}
\min \{G_{{\boldsymbol K}}({\boldsymbol K}_{\mathbf{pre}}(P_a),O{\boldsymbol K}_{\mathbf{pre}}(P_b))\}=arc\cos tr(\Lambda),
\label{e6}
\end{equation}
where $R$ is the identity matrix that can be regarded as the alignment based on the eigenvalues, and the trace of the $\Lambda$ corresponds to the minimum value of the measurement. The rotation $O_r$ and $G_{{\boldsymbol K}}$ for registration can be deduced as
\begin{equation}
{\boldsymbol K}_{\mathbf{pre}}(P_a)\cdot (O_r{\boldsymbol K}_{\mathbf{pre}}(P_b)) = tr(\Lambda), O\leftarrow O_r,
\label{e7}
\end{equation}
\begin{equation}
G_{{\boldsymbol K}}({\boldsymbol K}_{\mathbf{pre}}(P_a), {\boldsymbol K}_{\mathbf{pre}}(P_b))=arc\cos tr(\Lambda).
\label{e7_0}
\end{equation}

\begin{figure}
  \centering
  \includegraphics[width=\linewidth]{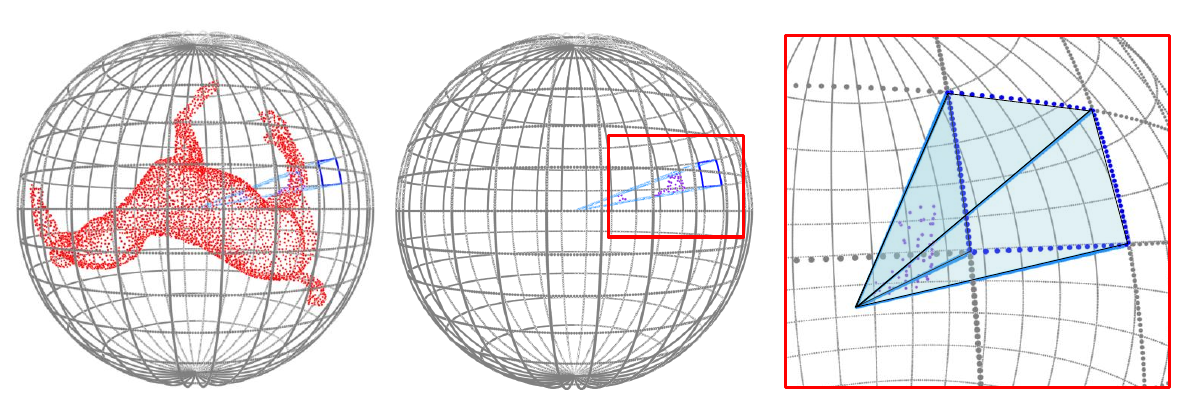}
  \vspace{-3mm}
  \caption{An instance of partition structure with sub-spaces.}
  \vspace{-3mm}
  \label{f3}
\end{figure}

Regrettably, the prerequisite for the above calculation is that point clouds should be ordered or have point-based correspondence according to the Wahba problem. In~\cite{lv2023kss}, the measurement for unorganized point cloud is simulated by Hausdorff distance which restricts the registration. To solve the problem, we provide a local shape matching method based on a partition structure that is similar to the 3D shape context~\cite{frome2004recognizing}. The point cloud is located into an inscribed ball space. The partition structure is a set of sub-spaces that are divided from the ball space. The sub-space is defined by the elevation and azimuth. In each sub-space, we select the point to be the representative sample that has largest distance to the center. Then, the local shape matching between points from different point clouds can be established based on the sub-spaces. The local shape matching can be represented as
\begin{equation}
\begin{array}{c}
\{p_1,...,p_i,...,p_k\}\in {\boldsymbol K}_{\mathbf{pre}}(P_a^{o}),\\ 
\{q_1,...,q_i,...,q_k\}\in {\boldsymbol K}_{\mathbf{pre}}(P_b^{o}),
\end{array}
\label{e7_1}
\end{equation}
where $p_i$ and $q_i$ are two representative samples from aligned sub-space $i$ of ${\boldsymbol K}_{\mathbf{pre}}(P_a^{o})$ and ${\boldsymbol K}_{\mathbf{pre}}(P_b^{o})$ according to same partition structure, $P_a^{o}$ and $P_b^{o}$ are subsets of $P_a$ and $P_b$, which represent the representative samples in $R^3$ space. In this way, $ G_{\boldsymbol K}$ can be computed by representative samples for current poses of two point clouds. In Fig.~\ref{f3}, we show an instance to explain the partition structure.

\begin{figure}
  \centering
  \includegraphics[width=\linewidth]{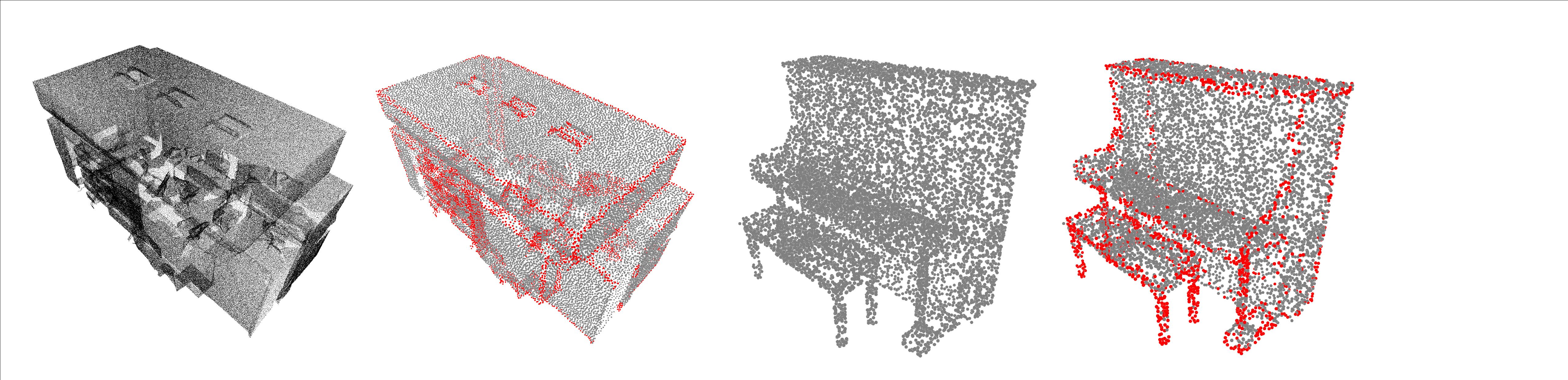}
  \vspace{-3mm}
  \caption{Instances of $P^{'}$ (red points). Internal geometric features are extracted according to $D_{pca}$.}
  \vspace{-3mm}
  \label{f3_1}
\end{figure}

Although the representative samples can be used to represent the external discrete shape form of a point cloud, the internal geometric features are ignored. It brings a potential disadvantage that the performance of measurement may deteriorate for point cloud with symmetrical outer contour. Such disadvantage has been discussed in KSS-ICP~\cite{lv2023kss}. To overcome the defect, we introduce geometric feature points to construct new representative sample set, which considers the internal geometry features for registration. Then the Eq~\eqref{e7_1} can be rewritten as
\begin{equation}
\begin{array}{c}
\{p_1^{'},...,p_i^{'},...,p_k^{'}\}\in {\boldsymbol K}_{\mathbf{pre}}(P_a^{'}),\\ 
\{q_1^{'},...,q_i^{'},...,q_k^{'}\}\in {\boldsymbol K}_{\mathbf{pre}}(P_b^{'}),
\end{array}
\label{e7_2}
\end{equation}
where $p_i^{'}$ and $q_i^{'}$ are geometric feature points, $P_a^{'}$ and $P_b^{'}$ are feature point sets. Based on the two representative sample sets, the Eq.~\eqref{e7_0} can be rewritten as
\begin{equation}\color{black}
\begin{array}{c}
G_{{\boldsymbol K}}({\boldsymbol K}_{\mathbf{pre}}(P_a), {\boldsymbol K}_{\mathbf{pre}}(P_b))=max\{G_{{\boldsymbol K}}({\boldsymbol K}_{\mathbf{pre}}(P_a^{o}), \\
{\boldsymbol K}_{\mathbf{pre}}(P_b^{o})), G_{{\boldsymbol K}}({\boldsymbol K}_{\mathbf{pre}}(P_a^{'}), {\boldsymbol K}_{\mathbf{pre}}(P_b^{'}))\},
\end{array}
\label{e7_3}
\end{equation}
where the new $G_{{\boldsymbol K}}$ is computed from the related subsets of original point clouds $P_a$ and $P_b$, which considers outer contour and internal geometric features. Even the extracted feature points are affected by defective parts and noise, $G_{{\boldsymbol K}}({\boldsymbol K}_{\mathbf{pre}}(P_a^{o}), {\boldsymbol K}_{\mathbf{pre}}(P_b^{o}))$ is used to keep the lower bound estimation (same to Eq.~\eqref{e7_0}). In practice, we employ a straightforward feature point extraction strategy that is based on the local fitting plane mapping distance $D_{pca}$, represented as
\begin{equation}\color{black}
P^{'}_a = \{p_i^{'}|D_{pca}(p_i)>d_{thr}, p_i\in P_a\},
\end{equation}
where $D_{pca}$ is the distance between point $p_i$ and its PCA-based fitting plane according to the $k$-neighbor region ($k=12$). If the point $p_i$ is located onto a plane, then the $D_{pca}(p_i) = 0$. On the contrary, the $D_{pca}(p_i)$ becomes larger when $p_i$ is located on the region with sharp curvature changing. It can be used to represent rough internal geometric features. We sort the $\{D_{pca}\}$ and set the threshold $d_{thr}$ according to the point number we need. In Fig.~\ref{f3_1}, instances of $P^{'}$ are visualized. 

As mentioned in Sec.~\ref{Pre-processing}, the non-uniform scaling should be estimated. The point set $\{x_1,...,x_m\}$ is changed to $\{x_1',...,x_k'\}$ that includes the representative samples based on the partition structure. \textcolor{black}{Based on the local correspondence, the scaling factor is recomputed by the Eq.~\eqref{e2}. Combined with the following global searching scheme, different scales can be normalized.} Even if the resampled point number is not strictly equal to $m$, the shape measurement is not affected. It should be noticed that the achieved $O_r$ and $ G_{\boldsymbol K}$ are just local registration results based on the current poses with local registration. To achieve the final registration result, the global searching strategy is required.

\begin{figure}
  \centering
  \includegraphics[width=\linewidth]{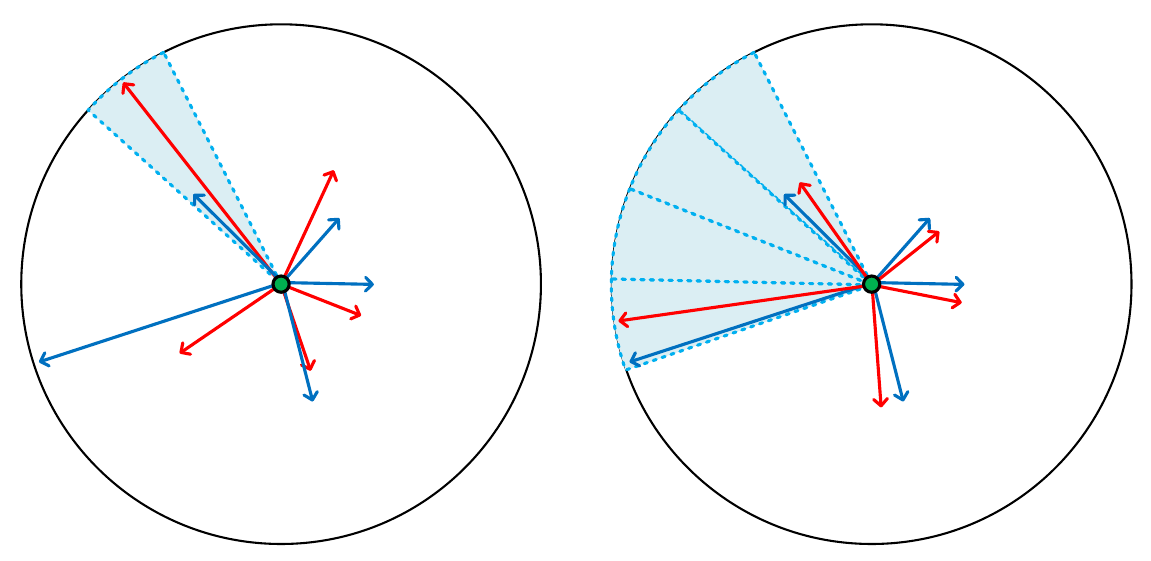}
  \vspace{-3mm}
  \caption{An instance of global searching based on partition structure. The red vector set is generated from representative samples and center from source point cloud, the blue vector set is generated from representative samples and center from template point cloud, the sector area labeled by light blue is the cell of the partition structure. The global searching is to find the final registration result in different cells one by one.}
  \vspace{-3mm}
  \label{f4}
\end{figure}

\section{Global Searching Scheme}\label{Global Searching Strategy}

The proposed shape measurement estimates the shape similarity between point clouds based on the current poses. It can be regarded as the local registration implemented in the partition structure. Naturally, the achieved rotation $O_r$ is only a local optimal result. To achieve the final registration result, we propose the global searching strategy to search the global optimum result in $SO(3)$.

Briefly, the global searching strategy is a controllable exhaustive searching process. Considering the shape measurement is implemented in the current pose, we just rotate the source point cloud according to the partition structure to change the pose. Then, the global searching in $SO(3)$ can be implemented. In Fig.~\ref{f4}, an instance is shown. Let $\{O_c\}$ to be a candidate rotation set, the target of the global searching strategy is to minimize the shape measurement based on the $\{O_c\}$, which can be represented as
\begin{equation}
arg\min({\boldsymbol K}_{\mathbf{pre}}(P_a)\cdot(O_rO_f{\boldsymbol K}_{\mathbf{pre}}(P_b))),O_f\in\{O_c\},
\label{e8}
\end{equation}
where $O_f$ is a rotation transformation of $\{O_c\}$. The global searching strategy is to find the $O_f$ to minimize shape measurement. The computational complexity of the strategy is proportional to the scale of $\{O_c\}$. Fortunately, the shape measurement with different rotation transformations can be implemented in parallel. We assign different computational units for $\{O_c\}$ to compute the shape measurement. An instance of GPU-based acceleration has been shown in Fig.~\ref{f2}. More implementation details are discussed in Appendix.

The prerequisite of the global searching strategy is that the centers of point clouds are aligned. However, the defective parts or missing sub-regions may change the semantic center of the point cloud. To solve the problem, we add a candidate translation set to fit the random movement of the center produced by the defective parts. The Eq~\eqref{e8} is updated as
\begin{equation}
\begin{array}{c}
arg\min({\boldsymbol K}_{\mathbf{pre}}(P_a)\cdot(O_rO_fS_i{\boldsymbol K}_{\mathbf{pre}}(P_b))),\\
O_f\in\{O_c\}, S_i\in\{S_c\},
\label{e9}
\end{array}
\end{equation}
where $\{S_c\}$ is the candidate translation set, $S_i$ is a translation matrix from the $\{S_c\}$ to represent the center movement. Combining the two candidate sets and shape measurement, the required transfer matrix can be represented by $O_r\times O_f\times S_i$. Then, the final transformation matrix can be estimated. It should be noticed that the candidate sets take influences for the efficiency and accuracy of the registration. we introduce the details of how to generate the candidate sets in practice. 

\begin{figure}
  \centering
  \includegraphics[width=\linewidth]{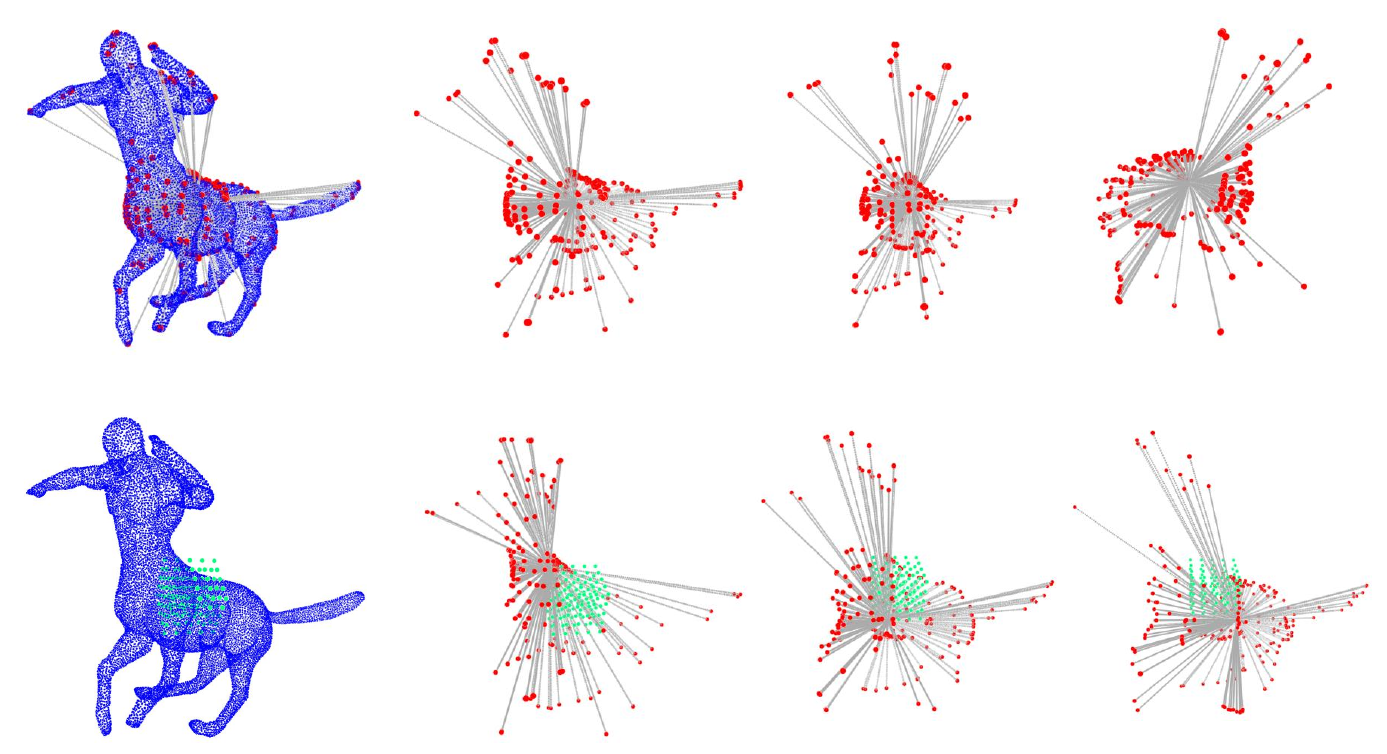}
  \vspace{-3mm}
  \caption{Instances of candidate rotation and translation sets. First row: instances of $O_c$; second row: instances of $S_c$ (green points represent candidate centers).}
  \vspace{-3mm}
  \label{f5}
\end{figure}

\begin{figure*}
  \centering
  \includegraphics[width=\linewidth]{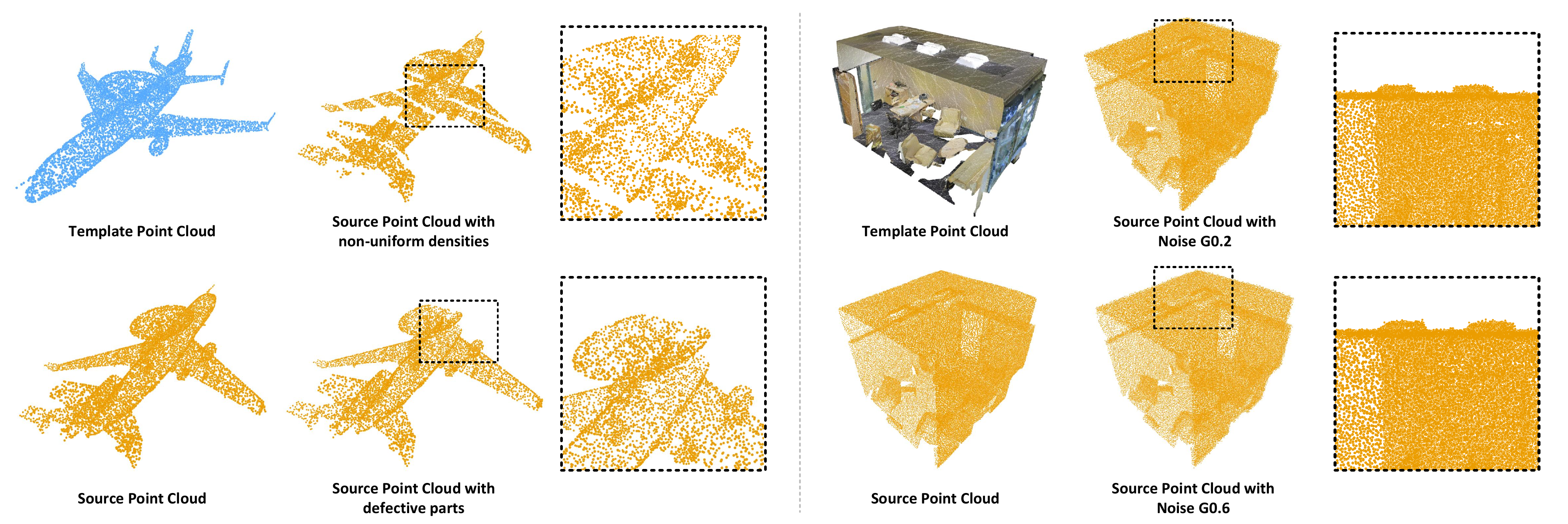}
  \vspace{-3mm}
  \caption{Generated source subsets with different types of interference factors, including non-uniform densities, defective parts, noisy points with different levels (G0.2 and G0,6 mean two kinds of Gaussian noisy distributions).}
  \vspace{-3mm}
  \label{f5_1}
\end{figure*}

For candidate rotation set $\{O_c\}$, we add the rotation units into the set according to the three axes by specified angle. It can be represented as
\begin{equation}
\begin{array}{c}
\{O_c\} = \{O_f|O_f=(n_x\theta_x, n_y\theta_y, n_z\theta_z),\\
n_x\theta_x, n_y\theta_y, n_z\theta_z \in\{0,2\pi\}\},
\end{array}
\label{e10}
\end{equation}
where $\theta_x, \theta_y$, and $\theta_z$ are unit angles based on three axes, $n_x, n_y$, and $n_z$ are specified steps to control the rotation. The parameters decide the scale of the candidate rotation set that should be consistent with the partition structure. In practice, we set $\theta_x=\theta_y=\theta_z=1/6\pi$, and the range of $n_x, n_y$, and $n_z$ is defined ($n_x, n_y, n_z\in[0,11]$). It can be computed that the candidate rotation set $O_c$ contains $12^3$ units. The number 12 is the rotation parameter.

\begin{algorithm}[t]
	\renewcommand{\algorithmicrequire}{\textbf{Input:}}
	\renewcommand{\algorithmicensure}{\textbf{Output:}}
	\caption{\textbf{Implementation of PKSS-Align}}
	\label{A1}
	\begin{algorithmic}[1]
        \REQUIRE Source and Template point clouds $P_s$ and $P_t$ 
        \ENSURE Aligned point cloud $P_{align}$
		\STATE Outliers pre-processing for $P_s$ and $P_t$ by Eq.~\eqref{e3} 
        \STATE Achieve ${\boldsymbol K}_{\mathbf p\mathbf r\mathbf e}(P_s)$ and ${\boldsymbol K}_{\mathbf p\mathbf r\mathbf e}(P_t)$ from processed $P_s$ and $P_t$ by Eq.~\eqref{e2} 
		\STATE Generate $\{O_c\}$ and $\{S_c\}$ by Eq.~\eqref{e10} and Eq.~\eqref{e11}
        \STATE Initialize $P_{c} \leftarrow P_s$
        \STATE Initialize $C_{c} \leftarrow G_k({\boldsymbol K}_{\mathbf{pre}}(P_t), {\boldsymbol K}_{\mathbf{pre}}(P_s))_{local}$ by Eq.~\eqref{e7_3}
	    \FOR{$(O_f,S_i) \in \{O_c\},\{S_c\}$}
           \STATE ${\boldsymbol K}_{\mathbf{pre}}(P_j)\leftarrow S_i\times O_f\times {\boldsymbol K}_{\mathbf{pre}}(P_s)$ 
           \STATE $C_j \leftarrow G_k({\boldsymbol K}_{\mathbf{pre}}(P_t), {\boldsymbol K}_{\mathbf{pre}}(P_j))_{local}$ by Eq.~\eqref{e7_3}          
           \IF{$C_j<C_{c}$}
               \STATE $P_{c}\leftarrow P_j$    
               \STATE $C_{c}\leftarrow C_j$ 
           \ENDIF
        \ENDFOR   
		\STATE $P_{align}\leftarrow P_{c}$   		
		\ENSURE $P_{align}$
	\end{algorithmic}  
\end{algorithm}

For candidate translation set $\{S_c\}$, we add different translations for the center to estimate the influences produced by defective parts. We generate the translations based on the local coordinate system that is established by principal component analysis (PCA). Following the PCA-based eigenvectors $u, v$, and $w$, we generate the translation units that represented as
\begin{equation}
\begin{array}{c}
\{S_c\} = \{S_i|S_i=(n_ul_u, n_vl_v, n_wl_w)\},
\end{array}
\label{e11}
\end{equation}
where $l_u, l_v$, and $l_w$ are unit vectors according to the eigenvectors, $n_u, n_v$, and $n_z$ are length factors that controls the translations. Unlike rotation set, there has not clear range for translations. We artificially define a boundary to control the scale of the translation set:
\begin{equation}
n_ul_u, n_vl_v, n_wl_w\in [-|u|/4, |u|/4],
\label{e12}
\end{equation}
where $|u|$ is the length of the eigenvector with largest eigenvalue. The range ensures the center movement can be captured when the defective parts is smaller than half according to experience. We select 5 units based on the range for each direction. Then, we achieve a translation set with $5^3$ units. The number 5 is the translation parameter. In Fig.~\ref{f5}, instances of $O_c$ and $S_c$ are shown. 

Based on the mentioned three components, the pipeline of PKSS-Align is constructed which can be concluded in Algorithm~\ref{A1}. It should be noticed that the rotation and translation parameters (12 and 5 by default) have significant impact for the performance of the registration. The larger values of them increase the computation cost without accuracy improvement for registration. Conversely, the candidate sets with insufficient rotations and center movements can not provide a functional global searching in $SO(3)$. The influences of the parameters with different values are evaluated in quantitative analysis of the experiment. 

\begin{figure*}
  \centering
  \includegraphics[width=\linewidth]{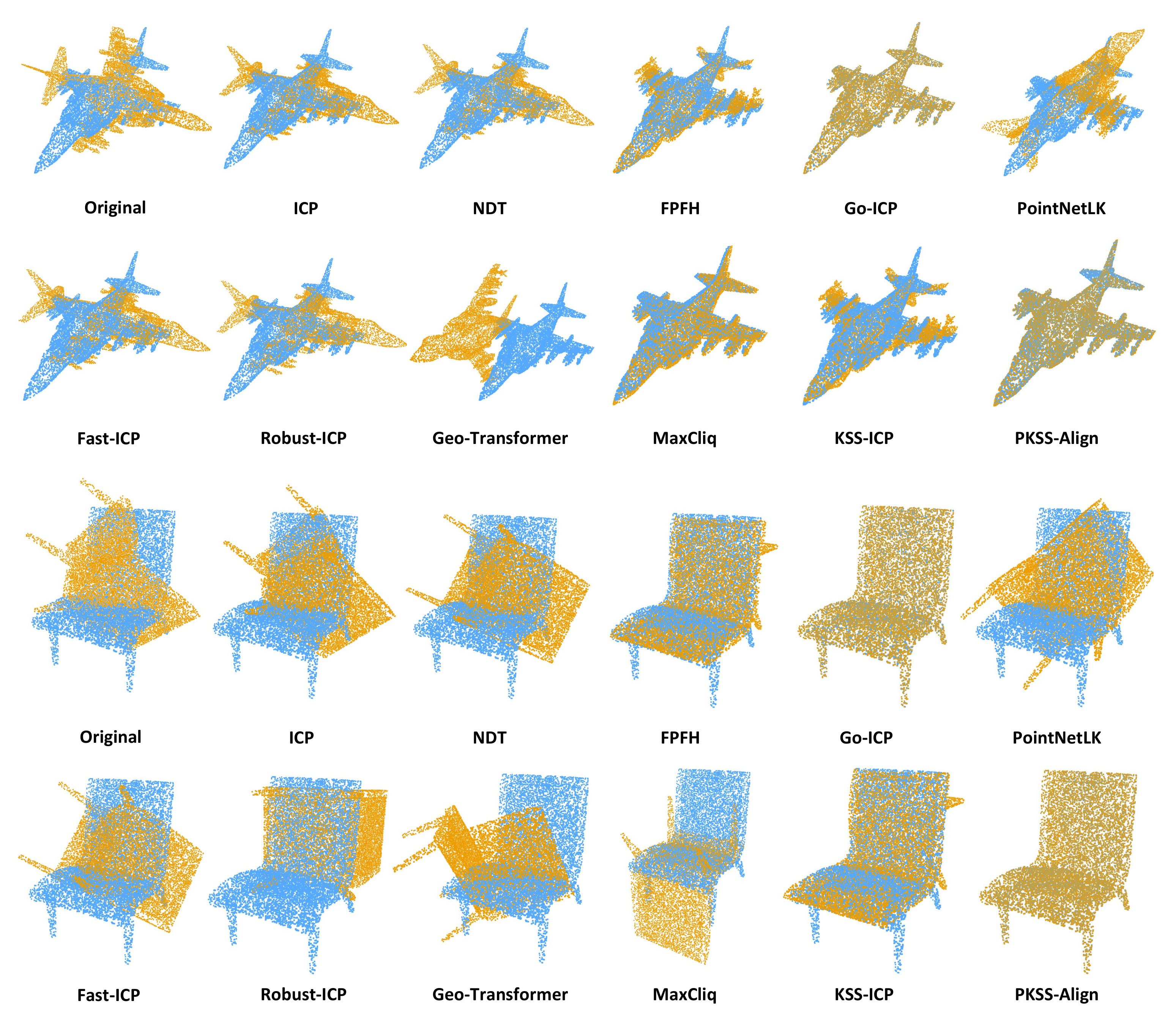}
  \vspace{-6mm}
  \caption{Registration results by different methods based on point clouds with similarity transformations.}
  \vspace{-4mm}
  \label{f6}
\end{figure*}

\section{Experiments}

We evaluate the performance of PKSS-Align in different datasets, including ModelNet40~\cite{wu20153d}, S3DIS dataset~\cite{armeni_cvpr16}, 3DMatch~\cite{zeng20173dmatch}, and KITTI~\cite{behley2019semantickitti}. The experimental machine equipped with Intel(R) i9-13900K 3.00 GHz, 128G RAM, GeForce RTX4090. The running system is Windows 11 with Visual Studio 2022 (64 bit) and Pycharm 2022 as the development platforms. The comparison methods involve various mainstream technical routes, including original ICP~\cite{besl1992method}, NDT~\cite{biber2003normal}, FPFH~\cite{yang2016fast}, Go-ICP~\cite{yang2013go}, PonitNetLK~\cite{aoki2019pointnetlk}, Fast and Robust-ICP~\cite{Zhang2022FastICP}, GeoTransformer~\cite{qin2022geometric}, Maximal
cliques~\cite{zhang20233d}, REGTR~\cite{yew2022regtr}, RoITr~\cite{yu2023rotation}, and KSS-ICP~\cite{lv2023kss}. Firstly, we introduce the selected datasets and related metrics for performance evaluation of registration. Secondly, we test the robustness of different methods for point clouds with different influence factors and report the quantitative results. Then, we evaluate our method on real scanning data. Finally, we show a comprehensive analysis of our method, including influences of parameter selections and algorithm characteristics analysis.

\begin{figure*}
  \centering
  \includegraphics[width=\linewidth]{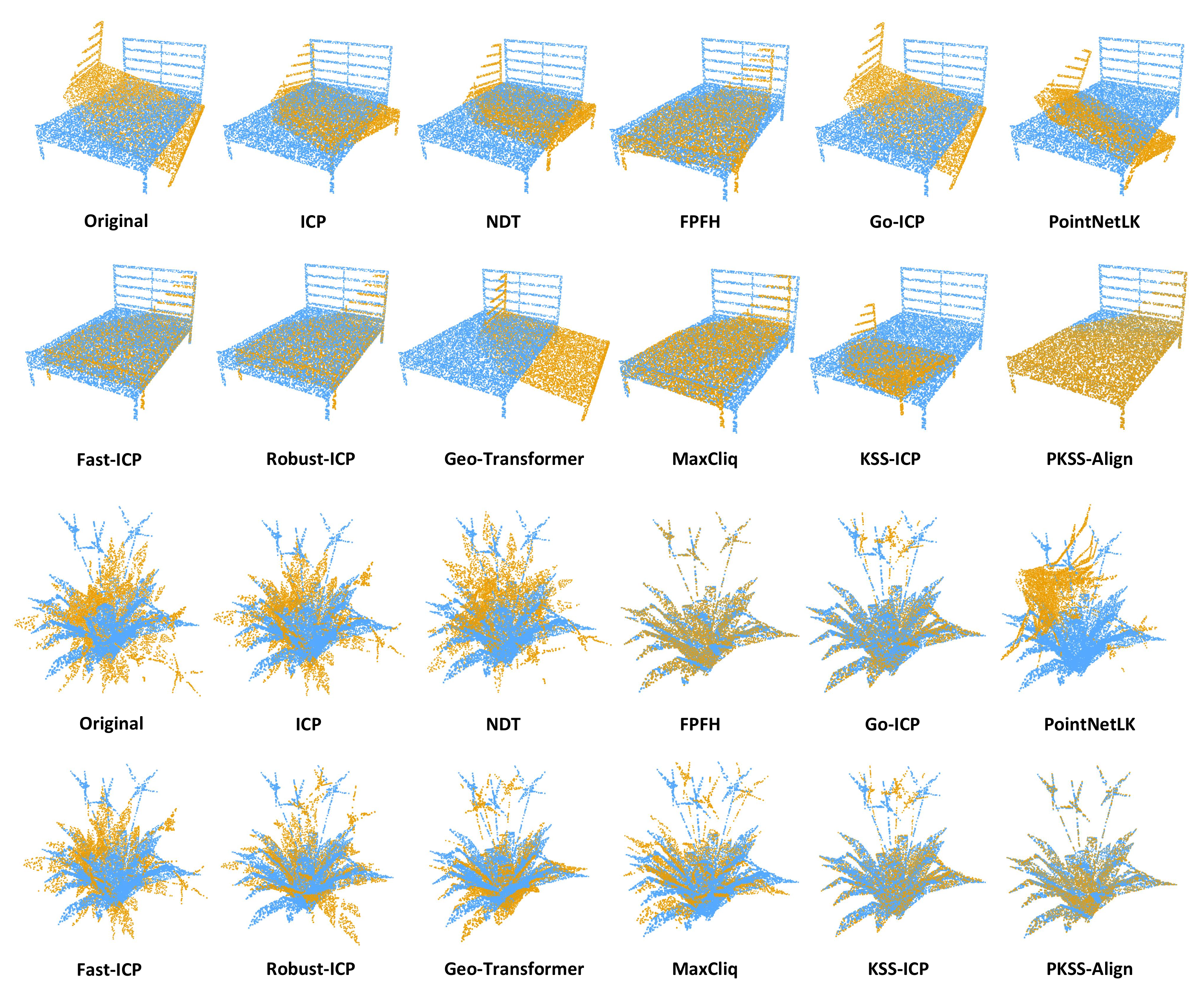}
  \caption{Registration results by different methods based on point clouds with similarity transformations, defective parts, and non-uniform densities.}
  \label{f7}
\end{figure*}

\subsection{Datasets and Metrics}

The test dataset of ModelNet40 contains 1235 models that is sampled from each category by 10\%. The S3DIS dataset contains 272 indoor scenes with multi-objects. The 3DMatch and KITTI datasets contains 171 models and 158 scenes. We add random similarity transformations into the datasets to generate related source datasets. The transformation matrices are recorded for further evaluation. The random similarity transformation includes random rotation $\theta\in[-\pi,\pi]$ and scaling $s\in[0.5, 2]$. For point clouds of S3DIS dataset, we implement the random resampling to achieve simplified set. The reason is that the data scale of original point cloud is too large ($>1,000K$), which limits the feasibility of some methods. To comprehensively evaluate the robustness of registration methods in various scenarios, we reconstruct a series of subsets with different types of interference factors based on test datasets, including point cloud with defective parts, non-uniform densities, and random noisy points with different levels. Some instances are shown in Fig.~\ref{f5_1}. Considering some registration methods can not handle the different scales of point clouds, a scale normalization method \cite{dimitrov2009bounds}(PCA-based bounding box) is used as the auxiliary function for comparative methods.

\begin{table*}[]\scriptsize
\centering
\caption{Evaluations in ModelNet40 and S3DIS datasets with similarity transformations.}
\label{t1}
\resizebox{\linewidth}{!}{
\begin{tabular}{l|ccccc|ccccc}
\toprule
   \textbf{Dataset}  & \multicolumn{5}{c|}{\textbf{ModelNet40}}                                             & \multicolumn{5}{c}{\textbf{S3DIS}}         \\ \midrule
   \textbf{Methods\textbackslash Metric}        & \textbf{Time}$\downarrow$   & \textbf{MSE}$\downarrow$    & \textbf{MSE(n)}$\downarrow$  & \textbf{${\boldsymbol {GT}}_{\boldsymbol cos}$}$\uparrow$ & \textbf{RR}$\uparrow$  & \textbf{Time}$\downarrow$    & \textbf{MSE}$\downarrow$     & \textbf{MSE(n)}$\downarrow$ & \textbf{${\boldsymbol {GT}}_{\boldsymbol cos}$}$\uparrow$ & \textbf{RR}$\uparrow$ \\ \midrule
\textbf{ICP}~\cite{besl1992method}   & 1.981s          & 0.01253           & 0.6915           & 0.4259  & 11\%       & 60.137s  & 0.7535  & 1.0017 &  0.2457   & $<$10\%       \\
\textbf{NDT}~\cite{biber2003normal}  & 3.719s          & 0.01157           & 0.6735           & 0.4079   & 10\%     & 19.591s  & 204.421 & 1.0886 &  0.2092   & $<$10\%         \\
\textbf{FPFH}~\cite{yang2016fast}  & 16.665s         & 0.00106          & 0.2163           & 0.6837  & 52\%     & 69.399s  & 0.06832 & 0.6871 &  0.3084   & 11\%         \\
\textbf{Go-ICP}~\cite{yang2013go}  & 33.101s         & \cellcolor{lightgray}{\textbf{0.00014}} & 0.0885         & 0.7824    & 68\%        & ——      & ——      & ——     & ——   & ——      \\
\textbf{PointNetLK}~\cite{aoki2019pointnetlk}  & 0.903s & 0.02621            & 0.7476           & 0.3871     & $<$10\%    & 7.334s  & 1.16521 & 0.9449 &  0.2852   & $<$10\%  \\
\textbf{Fast-ICP}~\cite{Zhang2022FastICP}  & 6.798s          & 0.00808          & 0.5255           & 0.4668  & 24\%        & 113.513s  & 0.08390 & 0.7673 &  0.3117   & 14\%     \\
\textbf{Robust-ICP}~\cite{Zhang2022FastICP}  & 22.921s        & 0.01207           & 0.4849           & 0.4838 & 26\%        & 251.649s  & 0.12021 & 0.7688 &  0.3177   & 13\%      \\
\textbf{Geo-Transformer}~\cite{qin2022geometric} &  \cellcolor{lightgray}{\textbf{0.563s}}             &       0.09309              &      0.8018              &       0.4776    & $<$10\%     & \cellcolor{lightgray}{\textcolor{black}{\textbf{0.987s}}}  & \textcolor{black}{2.3514}& \textcolor{black}{1.1223} &  \textcolor{black}{0.2292}   & \textcolor{black}{$<$10\%}  \\
\textbf{MaxCliq}~\cite{zhang20233d}  & 2.513s          & 0.02898           & 0.4558           & 0.6688   & 22\%        & ——  & —— & —— &  ——   & —— 
      \\ 
\textbf{REGTR}~\cite{yew2022regtr} & 8.8032s  & 0.37211 & 1.12591  & 0.2551 & $<$10\%  & —— & ——  & —— & —— &      —— \\
\textbf{RoITr}~\cite{yu2023rotation}   &  0.6283s &  0.00846  &  0.2472  & 0.6124 & 49\%   & —— & ——  & —— & ——  &  —— \\
      \midrule
\textbf{KSS-ICP}~\cite{lv2023kss}  & 2.899s          & 0.00033          & 0.1233           & 0.7648    & 65\%        & 11.746s  & 0.00837 & 0.1199 &  0.8604   & 79\%   \\
\textbf{PKSS-Align}   & 2.938s          & 0.00051          & \cellcolor{lightgray}{\textbf{0.0589}} & \cellcolor{lightgray}{\textbf{0.9026}} & \cellcolor{lightgray}{\textbf{80\%}} & 12.226s  & \cellcolor{lightgray}{\textbf{0.00493}} & \cellcolor{lightgray}{\textbf{0.1187}} &  \cellcolor{lightgray}{\textbf{0.8826}}   & \cellcolor{lightgray}{\textbf{82\%}}  
  \\\bottomrule                 
\end{tabular}}
\end{table*}

\begin{figure*}
  \centering
  \includegraphics[width=\linewidth]{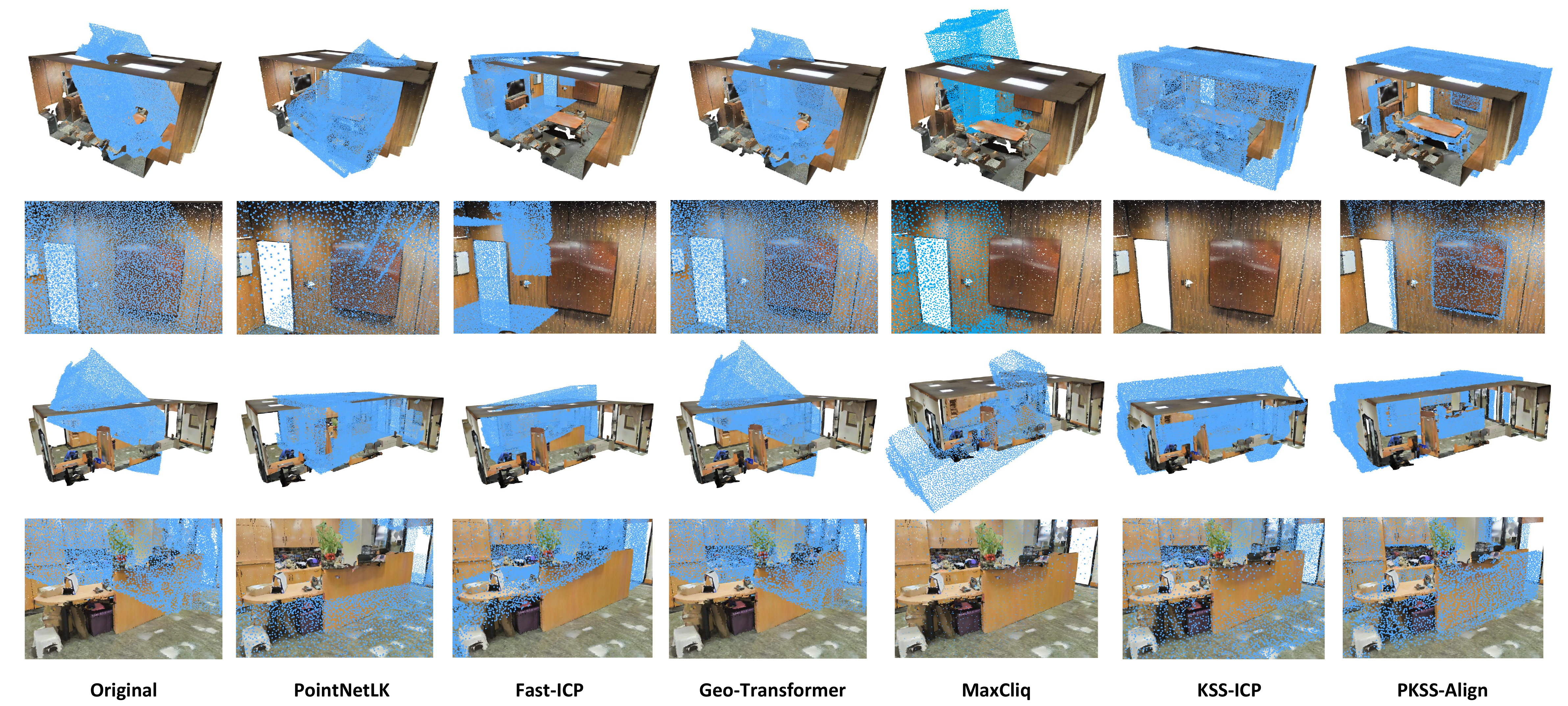}
  \caption{Registration results by different methods in the S3DIS with similarity transformations and different kinds of noise. First and second lines show point clouds with Gaussian noise (r = 0.6); third and fourth lines show point clouds with mean noise (r = 0.6).}
  \label{f8}
\end{figure*}

\begin{table*}[]\scriptsize
\centering
\caption{Evaluations in ModelNet40 dataset with similarity transformations, defective parts, and non-uniform densities.}
\label{t2}
\resizebox{\linewidth}{!}{
\begin{tabular}{l|ccccc|ccccc}
\toprule
   \textbf{Dataset}            & \multicolumn{5}{c|}{\textbf{ModelNet40 (defective parts)}}                                             & \multicolumn{5}{c}{\textbf{ModelNet40 (non-uniform densities)}}         \\ \midrule
   \textbf{Methods\textbackslash Metric}        & \textbf{Time}$\downarrow$     & \textbf{MSE} $\downarrow$     & \textbf{MSE(n)}$\downarrow$  & \textbf{${\boldsymbol {GT}}_{\boldsymbol cos}$}$\uparrow$ & \textbf{RR}$\uparrow$   & \textbf{Time}$\downarrow$    & \textbf{MSE}$\downarrow$     & \textbf{MSE(n)}$\downarrow$ & \textbf{${\boldsymbol {GT}}_{\boldsymbol cos}$}$\uparrow$ & \textbf{RR}$\uparrow$ \\ \midrule
\textbf{ICP}~\cite{besl1992method}   & 1.371s    & 0.01197    & 0.7832   & 0.3997 & 13\%    & 1.958s    & 0.01236   & 0.6973   & 0.4266 & 17\% \\
\textbf{NDT}~\cite{biber2003normal}  & 3.362s  & 0.01225   & 0.7717 & 0.3804 & 11\%   & 4.842s   & 0.01171  & 0.6806    & 0.4084 & 15\%     \\
\textbf{FPFH}~\cite{yang2016fast}   & 11.245s    & 0.00338     & 0.3923   & 0.6674 & 47\%  & 15.991s   & 0.00107   & 0.2381   & 0.7062 & 56\%   \\
\textbf{Go-ICP}~\cite{yang2013go}   & ——   & ——    & ——     & —— & ——   & 23.961   & 0.00014   & 0.1342    & 0.7811 & 67\%  \\
\textbf{PointNetLK}~\cite{aoki2019pointnetlk}  & 0.813s  & 0.03543   & 0.8505    & 0.3856 & $<$10\%  & 0.658s   & 0.02551   & 0.7569  & 0.3867 & 11\%  \\
\textbf{Fast-ICP}~\cite{Zhang2022FastICP}  & 6.235s  & 0.01261  & 0.5945    & 0.4536 & 20\%   & 5.625s   & 0.00794   & 0.539   & 0.4712 & 24\% \\
\textbf{Robust-ICP}~\cite{Zhang2022FastICP}  & 43.461  & 0.01243  & 0.5896    & 0.4556 & 21\%   & 30.637s  & 0.01179   & 0.5115   & 0.4801 & 26\% \\
\textbf{Geo-Transformer}~\cite{qin2022geometric} &  \cellcolor{lightgray}{\textbf{0.667s}} &	0.09329	&0.8423&	0.4415 & 13\%  &	\cellcolor{lightgray}{\textbf{0.628s}} &	0.0952	&0.8358&	0.4542 & 17\%
     \\
\textbf{MaxCliq}~\cite{zhang20233d}   & 1.236s    & 0.02359    & 0.4665    & 0.6856 & 45\%  & 1.212s     & 0.02318    & 0.4369     & 0.7016 & 51\%   \\ 
\textbf{REGTR}~\cite{yew2022regtr} & 7.764s  & 0.34431 & 1.1349  & 0.2481 & $<$10\%  & 7.895s & 0.37201  & 1.1249 & 0.2579  &  $<$10\% \\
\textbf{RoITr}~\cite{yu2023rotation}   & 0.689s  & 0.03012 & 0.5855  & 0.4885 & 32\%  & 0.652s  & 0.01539 & 0.3836  & 0.6938 & 54\%  \\
\midrule
\textbf{KSS-ICP}~\cite{lv2023kss}   & 39.438    & \cellcolor{lightgray}{\textbf{0.00152}}   & 0.4263    & 0.5886 & 40\%    & 2.697s    & 0.00039    & 0.1753    & 0.7677 & 65\%  \\
\textbf{PKSS-Align}  & 3.652s    & 0.00712    & \cellcolor{lightgray}{\textbf{0.3636}}   & \cellcolor{lightgray}{\textbf{0.7039}} & \cellcolor{lightgray}{\textbf{57\%}}   & 2.901s    & \cellcolor{lightgray}{\textbf{0.00013}}   & \cellcolor{lightgray}{\textbf{0.1041}}   & \cellcolor{lightgray}{\textbf{0.9115}} & \cellcolor{lightgray}{\textbf{87\%}} \\\bottomrule                 
\end{tabular}}
\end{table*}

\begin{table*}[]\scriptsize
\centering
\caption{Evaluations in S3DIS dataset with similarity transformations, defective parts, and non-uniform densities.}
\label{t2_2}
\resizebox{\linewidth}{!}{
\begin{tabular}{l|ccccc|ccccc}
\toprule
   \textbf{Dataset}            & \multicolumn{5}{c|}{\textbf{S3DIS (defective parts)}}                                             & \multicolumn{5}{c}{\textbf{S3DIS (non-uniform densities)}}         \\ \midrule
   \textbf{Methods\textbackslash Metric}        & \textbf{Time}$\downarrow$     & \textbf{MSE} $\downarrow$     & \textbf{MSE(n)}$\downarrow$  & \textbf{${\boldsymbol {GT}}_{\boldsymbol cos}$}$\uparrow$ & \textbf{RR}$\uparrow$   & \textbf{Time}$\downarrow$    & \textbf{MSE}$\downarrow$     & \textbf{MSE(n)}$\downarrow$ & \textbf{${\boldsymbol {GT}}_{\boldsymbol cos}$}$\uparrow$ & \textbf{RR}$\uparrow$ \\ \midrule
\textbf{ICP}~\cite{besl1992method}   & 51.821s  & 1.5664  & 1.0224 &  0.2747   & 27\%    & 44.852s    & 1.2599   & 1.0003   & 0.2865   & 11\%  \\
\textbf{NDT}~\cite{biber2003normal}  & 24.119s  & 204.103 & 1.1067 & 0.2095    & $<$10\%  & 13.012s    & 204.401   & 1.0851   & 0.2326 & $<$10\%    \\
\textbf{FPFH}~\cite{yang2016fast}   & 41.895s  & 0.1414  & 0.7553 &  0.3955   & 11\%    & 37.119s   & 0.07430   & 0.7014   & 0.4719 &  25\%  \\
\textbf{PointNetLK}~\cite{aoki2019pointnetlk}  &    5.497s &	1.5134 &	0.9598 &	0.3071 & $<$10\%  & 6.253s   & 1.1028   & 0.9399   & 0.3224 & $<$10\% \\
\textbf{Fast-ICP}~\cite{Zhang2022FastICP}  & 70.096s &   0.1131	&  0.7594  & 0.3656  &  11\%  & 44.302s   & 0.08438   & 0.7631   & 0.3093 & $<$10\%  \\
\textbf{Robust-ICP}~\cite{Zhang2022FastICP}   & ——      & ——      & ——     & ——   & ——   & 213.719s    & 0.1156   & 0.7627   & 0.3276 &  12\% \\
\textbf{Geo-Transformer}~\cite{qin2022geometric} &    \cellcolor{lightgray}{\textbf{0.987s}}	 &  2.3514  &	1.1223 &	0.2292 & $<$10\%   & \cellcolor{lightgray}{\textbf{1.036s}}    & 2.235   & 1.0968   & 0.2242 & $<$10\%   \\
\textbf{MaxCliq}~\cite{zhang20233d}  &    15.223s &	2.4223 &	0.9368	& 0.2833 & $<$10\%   &——    &——  &——  &—— &——  \\ 
\midrule
\textbf{KSS-ICP}~\cite{lv2023kss}   &   14.298s	& 0.0881 &	0.6033 &	0.4588 & 22\%   & 8.466s   & 0.008371   & \cellcolor{lightgray}{\textbf{0.1350}}   & 0.8583 & 78\% \\
\textbf{PKSS-Align}  &   12.961s &	\cellcolor{lightgray}{\textbf{0.0527}} &	\cellcolor{lightgray}{\textbf{0.4744}} & \cellcolor{lightgray}{\textbf{0.6751}} & \cellcolor{lightgray}{\textbf{49\%}} & 14.321s & \cellcolor{lightgray}{\textbf{0.007019}} & 0.1409 & \cellcolor{lightgray}{\textbf{0.8699}} & \cellcolor{lightgray}{\textbf{80\%}}     \\\bottomrule                 
\end{tabular}}
\end{table*}

\begin{table*}[]
\centering
\caption{Evaluations in S3DIS dataset with different kinds of Gaussian noise.}
\vspace{-2mm}
\label{t3}
\resizebox{\linewidth}{!}{
\begin{tabular}{l|cccc|cccc|cccc}
\toprule
\textbf{Dataset}        & \multicolumn{4}{c|}{\textbf{S3DIS (r = 0.2)}}                                   & \multicolumn{4}{c|}{\textbf{S3DIS (r = 0.4)}}                                   & \multicolumn{4}{c}{\textbf{S3DIS (r = 0.6)}}                                   \\ \midrule
\textbf{Methods\textbackslash Metric}      & \textbf{Time}$\downarrow$     & \textbf{MSE} $\downarrow$     & \textbf{MSE(n)}$\downarrow$  & \textbf{${\boldsymbol {GT}}_{\boldsymbol cos}$}$\uparrow$   & \textbf{Time}$\downarrow$    & \textbf{MSE}$\downarrow$     & \textbf{MSE(n)}$\downarrow$ & \textbf{${\boldsymbol {GT}}_{\boldsymbol cos}$}$\uparrow$ & \textbf{Time}$\downarrow$     & \textbf{MSE} $\downarrow$     & \textbf{MSE(n)}$\downarrow$  & \textbf{${\boldsymbol {GT}}_{\boldsymbol cos}$}$\uparrow$ \\ \midrule
\textbf{ICP}~\cite{besl1992method}      & 74.333s       & 1.4551       & 1.0124          &       0.2736     & 65.401s       & 1.532        & 1.021           &      0.2718              & 62.572s        & 1.5216       & 1.023           &      0.2719               \\
\textbf{NDT}~\cite{biber2003normal}      & 20.843s       & 182.228      & 1.10314         &       0.2131               & 22.098s        & 204.068      & 1.1038          &        0.2102     & 23.578s       & 204.103      & 1.1044          &      0.2095              \\
\textbf{FPFH}~\cite{yang2016fast}     & 44.822s       & 0.1154       & 0.7374          &        0.4392            & 46.595s       & 0.1223       & 0.7442          &        0.4125          & 46.303s       & 0.1234       & 0.7399          &      0.3989           \\
\textbf{PointNetLK}~\cite{aoki2019pointnetlk}     &   8.555s &	1.4534&	0.9649&	0.3055&	6.316s&	1.5078 &	0.9644 &	0.3051 &	5.739s &	1.5004 &	0.9638 &	0.3052
       \\
\textbf{Fast-ICP}~\cite{Zhang2022FastICP}    &  80.642s &	0.1104 &	0.7673 &	0.3537 &	90.602s &	0.1098 &	0.7674 &	0.3477 &	118.256s &	0.1112 &	0.7751 &	0.3617\\

\textbf{Geo-Transformer}~\cite{qin2022geometric}     &   \cellcolor{lightgray}{\textbf{ 1.236s}}	&2.2317&	1.1239&	0.2272&	\cellcolor{lightgray}{\textbf{1.243s}}&	2.1828&	1.1177&	0.2251&	\cellcolor{lightgray}{\textbf{1.251s}}&	2.3735&	1.1271&	0.2199
        \\
\textbf{MaxCliq}~\cite{zhang20233d}  &      17.172s &	2.0971 &	0.9671 &	0.2609 &	20.931s &	2.0537 &	0.9826&	0.2856 &	23.041s &	2.2453 &	0.9625 &	0.2741
      \\       \midrule
\textbf{KSS-ICP}~\cite{lv2023kss}   &    48.242s  & 	0.08742  & 	0.6105  & 	0.4661  & 	47.229s  & 	0.08299  & 	0.5996  & 	0.4704  & 	45.842s  & 	0.09519  & 	0.6131  & 	0.4568
           \\
\textbf{PKSS-Align} &     15.928s&	\cellcolor{lightgray}{\textbf{0.05037}}&	\cellcolor{lightgray}{\textbf{0.4802}}&	\cellcolor{lightgray}{\textbf{0.6661}}&	13.347s&	\cellcolor{lightgray}{\textbf{0.05172}}&	\cellcolor{lightgray}{\textbf{0.4814}}&	\cellcolor{lightgray}{\textbf{0.6798}}&	14.241s&	\cellcolor{lightgray}{\textbf{0.05172}}&	\cellcolor{lightgray}{\textbf{0.4814}} &	\cellcolor{lightgray}{\textbf{0.6705}}
            \\  \bottomrule               
\end{tabular}}
\vspace{-2mm}
\end{table*}

\begin{table*}[]
\centering
\caption{Evaluations in S3DIS dataset with different kinds of mean noise.}
\vspace{-2mm}
\label{t4}
\resizebox{\linewidth}{!}{
\begin{tabular}{l|cccc|cccc|cccc}
\toprule
\textbf{Dataset}        & \multicolumn{4}{c|}{\textbf{S3DIS (r = 0.2)}}                                   & \multicolumn{4}{c|}{\textbf{S3DIS (r = 0.4)}}                                   & \multicolumn{4}{c}{\textbf{S3DIS (r = 0.6)}}                                   \\ \midrule
\textbf{Methods\textbackslash Metric}      & \textbf{Time}$\downarrow$     & \textbf{MSE} $\downarrow$     & \textbf{MSE(n)}$\downarrow$  & \textbf{${\boldsymbol {GT}}_{\boldsymbol cos}$}$\uparrow$   & \textbf{Time}$\downarrow$    & \textbf{MSE}$\downarrow$     & \textbf{MSE(n)}$\downarrow$ & \textbf{${\boldsymbol {GT}}_{\boldsymbol cos}$}$\uparrow$ & \textbf{Time}$\downarrow$     & \textbf{MSE} $\downarrow$     & \textbf{MSE(n)}$\downarrow$  & \textbf{${\boldsymbol {GT}}_{\boldsymbol cos}$}$\uparrow$ \\ \midrule
\textbf{ICP}~\cite{besl1992method}    & 51.618s &	1.4901&	1.0256&	0.2692&	60.464s&	1.5154&	1.0173&	0.2744&	60.438s&	1.5048&	1.0277 &	0.2698   \\
\textbf{NDT}~\cite{biber2003normal}      & 19.339s	& 204.103 &	1.1047&	0.2105&	21.126s&	204.151&	1.1061&	0.2089&	22.675s&	204.07&	1.1046&	0.2103  \\
\textbf{FPFH}~\cite{yang2016fast}     & 41.247s &	0.1259&	0.7456&	0.4073&	43.105s&	0.1212&	0.7378&	0.4121&	45.351s&	0.1191&	0.7195&	0.4382   \\
\textbf{PointNetLK}~\cite{aoki2019pointnetlk}     &     5.871s&	1.5422&	0.9613&	0.305&	7.902s&	1.5354&	0.9636&	0.3047&	5.735s&	1.5537&	0.9646&	0.3046   \\
\textbf{Fast-ICP}~\cite{Zhang2022FastICP}     &      130.751s&	0.1141&	0.7616&	0.3528&	93.272s&	0.1113&	0.7608&	0.3641&	87.438s&	0.1135&	0.7711&	0.3668   \\
\textbf{Geo-Transformer}~\cite{qin2022geometric}    &   \cellcolor{lightgray}{\textbf{ 1.102s}}	 &1.9686 &	1.1161 &	0.2331 &	\cellcolor{lightgray}{\textbf{1.111s}} &	2.4561 &	1.1273 &	0.2383 &	\cellcolor{lightgray}{\textbf{1.208s}} &	2.4974 &	1.1211 &	0.2288
      \\
\textbf{MaxCliq}~\cite{zhang20233d}  &      12.441s &	2.5799 &	0.8989 &	0.2878 &	14.252s &	2.2542 &	0.9224 &	0.2661 &	17.142s &	2.2061 &	0.9751 &	0.2706
      \\  \midrule
\textbf{KSS-ICP}~\cite{lv2023kss}   &      46.989s & 0.0980&	0.6059&	0.4669&	46.915s&	0.0879&	0.6061&	0.4633&	45.151s&	0.0816 &	0.6024 &	0.4762
    \\
\textbf{PKSS-Align} &  11.836s & \cellcolor{lightgray}{\textbf{0.0526}} & \cellcolor{lightgray}{\textbf{0.4761}} &	\cellcolor{lightgray}{\textbf{0.6581}} &	12.698s &	\cellcolor{lightgray}{\textbf{0.0518}} &	\cellcolor{lightgray}{\textbf{0.4836}} &	\cellcolor{lightgray}{\textbf{0.6399}} &	12.915s &	\cellcolor{lightgray}{\textbf{0.0501}} &	\cellcolor{lightgray}{\textbf{0.4753}} &	\cellcolor{lightgray}{\textbf{0.6753}}
           \\  \bottomrule               
\end{tabular}}
\vspace{-2mm}
\end{table*}

To estimate the accuracy of the registration, we introduce two mean squared errors (MSE) to provide qualitative results, including closest point MSE and normal-based MSE(n). The closest point MSE reflects the overlap degree intuitively. The normal-based MSE(n) measures the local shape similarity based on normal vector angle. Since we record the transformation matrices for generation of source dataset, such matrices are used as ground truth to support a similarity measurement for registration. We employ the cosine similarity to be the measurement represented as
\begin{equation}
GT_{cos} = arccos\left\langle T_g,T^{'}\right\rangle,
\end{equation}
where $T_g$ represent the recorded rotation matrix as the ground truth with $3\times 3$ values, $T^{'}$ is the generated rotation matrix as the registration result, the cosine similarity $GT_{cos}$ and registration recall (RR) can be used as the quantitative evaluations for the quality of registration. We set $GT_{cos}>0.8$ and MSE$< 0.001$ (MSE-based condition is not used for 3DMatch and KITTI datasets, sparse densities take unstable values) as the successful conditions to obtain the statistical values of RR. In following parts, we compare the performance of different registration methods.

\textcolor{black}{\textbf{Implementation.} For the GPU-based acceleration, we parallel to compute the shape measurement between the representations with $12^3\times5^3$ ($\{O_c\}\times\{S_c\}$) times. In kernel function, the shape measurement requires allocating some GPU memory to store the intermediate results. Therefore, the number of threads must be controlled to avoid GPU memory overflow. In practice, we set the thread number to be 288 ($12^2\times2$) for 750 blocks ($12^3\times5^3/288$).} \textcolor{black}{For \textbf{ICP}~\cite{besl1992method}, the basic iteration step is set to 20, and the transformation $\epsilon$ is set to $0.1$. For \textbf{NDT}~\cite{biber2003normal}, the random sample is set to 1,000, the transformation $\epsilon$ is set to 0.01, the step size is set to 0.05, the resolution is set to 3. For \textbf{FPFH}~\cite{yang2016fast}, the random sample is set to 2,000, the searching radius is set to 0.05. For \textbf{Geo-Transformer}~\cite{qin2022geometric}, pre-training weights are selected from 3DMatch. Since the model of Geo-Transformer is sensitive to the scale, we change the initial voxel size ($0.025\rightarrow 0.1$) to fit the data of S3DIS. For \textbf{MaxCliq}~\cite{zhang20233d}, we control the maximum of maximal clique ($<10k$) to avoid non-convergence situation. For \textbf{KSS-ICP}~\cite{lv2023kss}, the rotation searching range is set to 12 that is same to PKSS-Align. The implementation codes of such methods are provided by original authors.} 

\subsection{Evaluation}

\textbf{Robustness to similarity transformations.} We evaluate the robustness of similarity transformations for different registration methods in the test datasets. In Fig.~\ref{f6}, some registration results are shown. The proposed PKSS-Align achieves better alignment between source point clouds and template ones. To further demonstrate the advantage of our approach in performance, we report the quantitative analysis in Table~\ref{t1}, which contains average time cost, MSE, MSE(n) and $GT_{cos}$. It is clear that our method achieves obviously improvement with SOTA methods in different kinds of models. \textcolor{black}{For RoITr, processing dense point clouds requires significant GPU memory overhead. Excessive down-sampling can lead to severe performance degradation in the algorithm. Its results on S3DIS are not included in the relevant tables. For REGTR, its convergence speed has exceeded the normal range ($>1$ hour), so some related data is also excluded from the tables.} Since the PCA-based scale normalization is used to uniform scales of point clouds, the traditional global searching strategy like Go-ICP can be converged efficiently. However, point clouds of S3DIS dataset have been simplified by random sampling, their global structures have been changed. It reduces the performance of scale normalization which affects the convergence of Go-ICP and Robust-ICP. Our method doesn't require the scale normalization while achieving the better alignment results. In addition, even PKSS-Align has similar structure with KSS-ICP, the proposed PKSS-based shape measurement provides more accurate metric for point cloud alignment. It supports more efficient point cloud alignment. 

\begin{figure*}
  \centering
  \includegraphics[width=\linewidth]{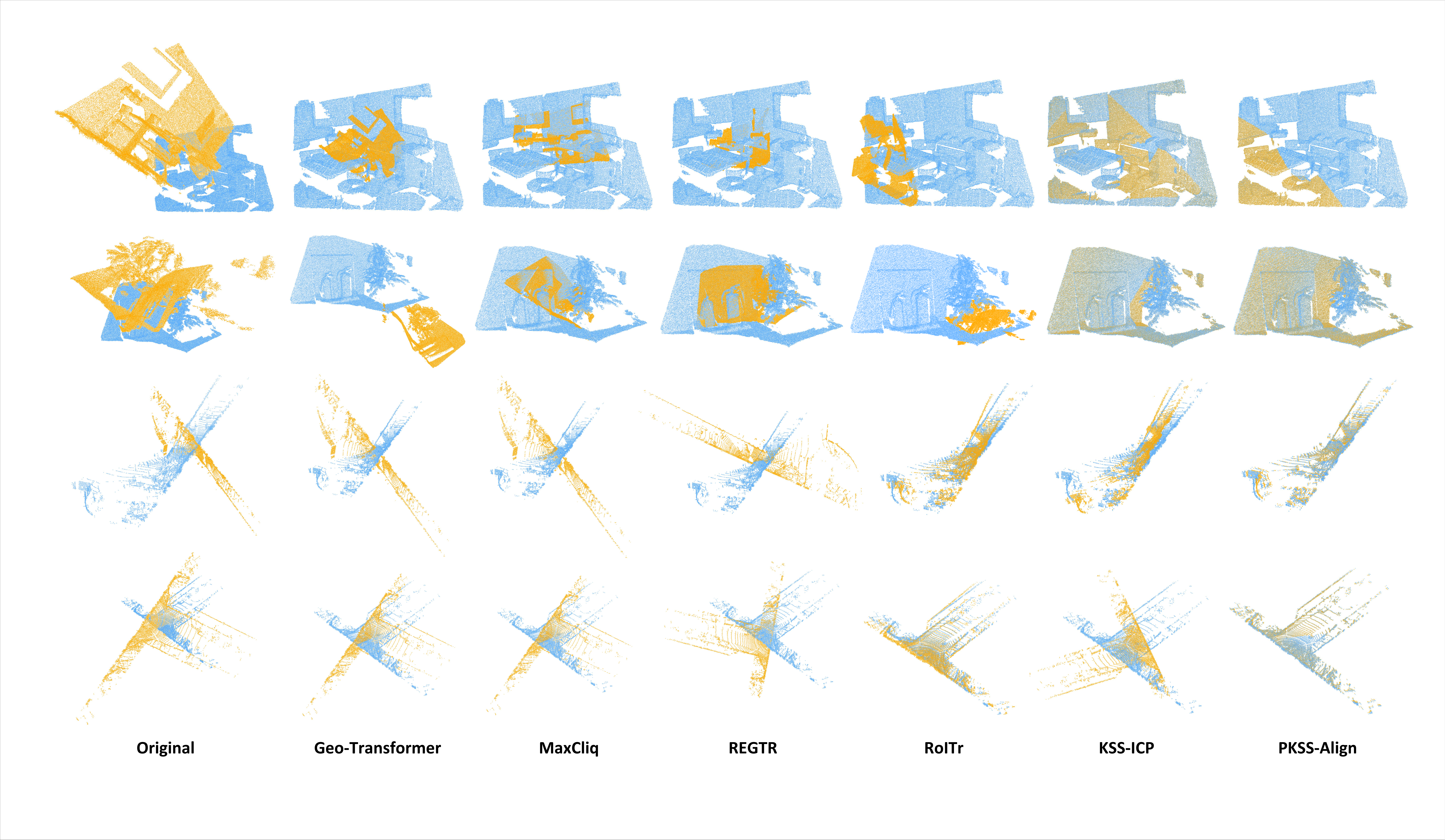}
  \vspace{-2mm}
  \caption{Registration results by SOTA methods in the 3DMatch and KITTI models with  significant posture differences.}
  \vspace{-2mm}
  \label{f8_0_0}
\end{figure*}

\begin{table*}[]\scriptsize
\centering
\caption{Evaluations in 3DMatch and KITTI datasets with similarity transformations.}
\label{t2_3}
\resizebox{\linewidth}{!}{
\begin{tabular}{l|ccccc|ccccc}
\toprule
   \textbf{Dataset}            & \multicolumn{5}{c|}{\textbf{3DMatch}}                                             & \multicolumn{5}{c}{\textbf{KITTI}}         \\ \midrule
   \textbf{Methods\textbackslash Metric}        & \textbf{Time}$\downarrow$     & \textbf{MSE} $\downarrow$     & \textbf{MSE(n)}$\downarrow$  & \textbf{${\boldsymbol {GT}}_{\boldsymbol cos}$}$\uparrow$ & \textbf{RR}$\uparrow$   & \textbf{Time}$\downarrow$    & \textbf{MSE}$\downarrow$     & \textbf{MSE(n)}$\downarrow$ & \textbf{${\boldsymbol {GT}}_{\boldsymbol cos}$}$\uparrow$ & \textbf{RR}$\uparrow$ \\ \midrule
\textbf{ICP}~\cite{besl1992method}   & 19.894s    & 0.02729    & 1.0244   & 0.3486 & 23\%    & 1.723s    & 165.131   & 0.9779   & 0.4915 & 33\% \\
\textbf{NDT}~\cite{biber2003normal}  & 19.919s    & 0.02736    & 1.0212   & 0.3528 & 18\%    & 2.616s    & 224.375   & 1.0595   & 0.3959 & 19\%  \\
\textbf{FPFH}~\cite{yang2016fast}   & 26.533s    & 0.00528    & 0.6322   & 0.8619  & 83\%    & 20.112s    & \cellcolor{lightgray}{\textbf{2.1131}}   & 0.6918   & \cellcolor{lightgray}{\textbf{0.9312}} & \cellcolor{lightgray}{\textbf{89\%}}  \\
\textbf{Go-ICP}~\cite{yang2013go}  & ——            & ——           & ——              & ——                 & ——    & 262.097s    & 6.5074   & 0.5948   & 0.6699 & 51\%  \\
\textbf{PointNetLK}~\cite{aoki2019pointnetlk}  & 2.591s    &  0.07212    & 1.0795   & 0.3491 & 16\%    & 0.643s    & 211.911   & 0.9963   & 0.4123  & 21\%  \\
\textbf{Fast-ICP}~\cite{Zhang2022FastICP}  & 39.856s    & 0.01926    & 0.9678   & 0.4071 & 24\%    & 3.204s    & 10.0256   & 0.7241   & 0.5560 & 35\% \\
\textbf{Robust-ICP}~\cite{Zhang2022FastICP}  & 269.319s  & 0.02557   & 0.9448   & 0.4186 & 24\%  & 25.301s    & 17.0675   & 0.6976   & 0.5548 & 34\% \\
\textbf{Geo-Transformer}~\cite{qin2022geometric} & \cellcolor{lightgray}{\textbf{0.689s}}  & 0.1921  & 1.1694   & 0.2329 & $<$10\%  & \cellcolor{lightgray}{\textbf{0.238s}}  & 411.909 & 1.2244  & 0.2204 & $<$10\%
     \\
\textbf{MaxCliq}~\cite{zhang20233d}   & 2.223s    & 0.05279    & 0.7652   & 0.8135 & 79\%    & 4.414s   & 398.298   & 1.1944   & 0.2366 & $<$10\%    \\ 
\textbf{REGTR}~\cite{yew2022regtr} & 86.848s  & 0.09001 & 0.9196  & 0.5456 & 44\%   & 61.1265s & 397.374  & 1.1682 & 0.2284  &  $<$10\% \\
\textbf{RoITr}~\cite{yu2023rotation}   & 3.8771s  & 0.06951 & 0.7578  & 0.7421 & 63\%  & 0.6898s & 77.299  & 0.8519 & 0.8297  &  79\% \\
\midrule
\textbf{KSS-ICP}~\cite{lv2023kss}   & 3.138s    & \cellcolor{lightgray}{\textbf{0.0005}}    & 0.0623  & 0.9485 & 93\%    & 9.776s    & 29.506   & 0.7474   & 0.5439 & 34\% \\
\textbf{PKSS-Align}  & 5.294s & 0.0008 & \cellcolor{lightgray}{\textbf{0.0398}}   & \cellcolor{lightgray}{\textbf{0.9785}} & \cellcolor{lightgray}{\textbf{97\%}} & 1.626s & 3.3382 & \cellcolor{lightgray}{\textbf{0.5113}} & 0.9063 & 87\% \\\bottomrule                 
\end{tabular}}
\end{table*}

\begin{figure*}
  \centering
  \includegraphics[width=\linewidth]{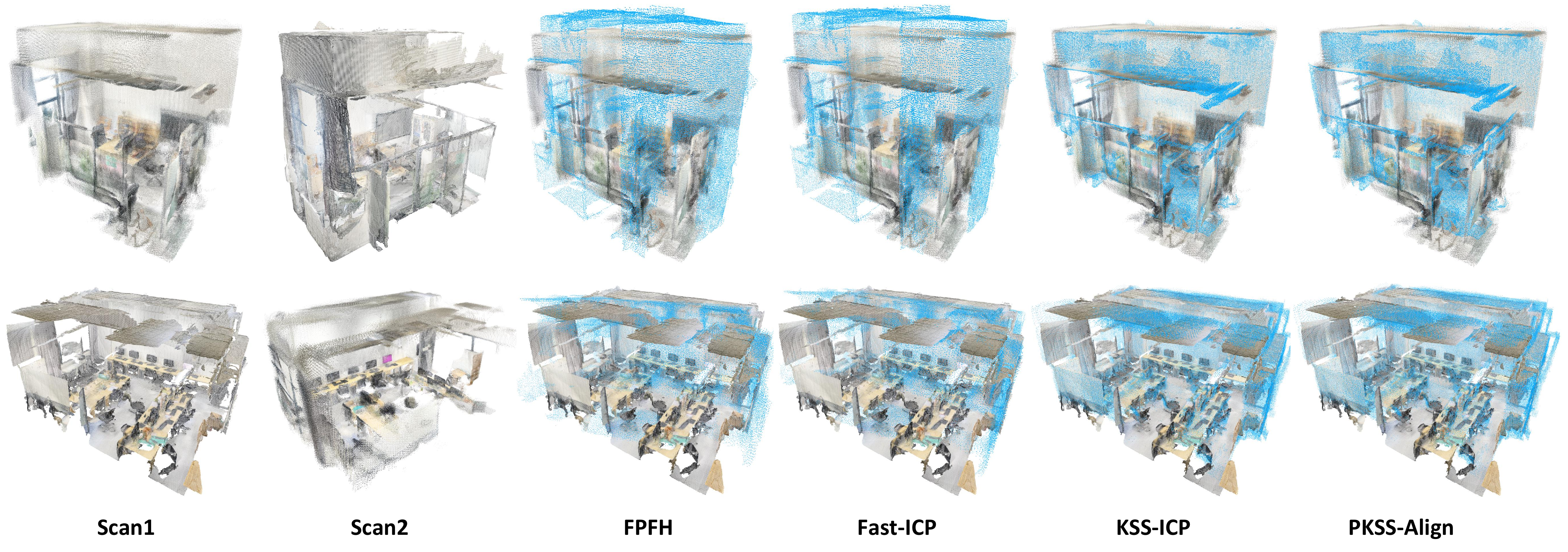}
  \vspace{-3mm}
  \caption{Registration results in the real scene. The registration methods align scan2 to scan1. Point clouds are scanned by iphone12pro.}
  \vspace{-3mm}
  \label{f8_0}
\end{figure*}

\begin{figure*}
  \centering
  \includegraphics[width=\linewidth]{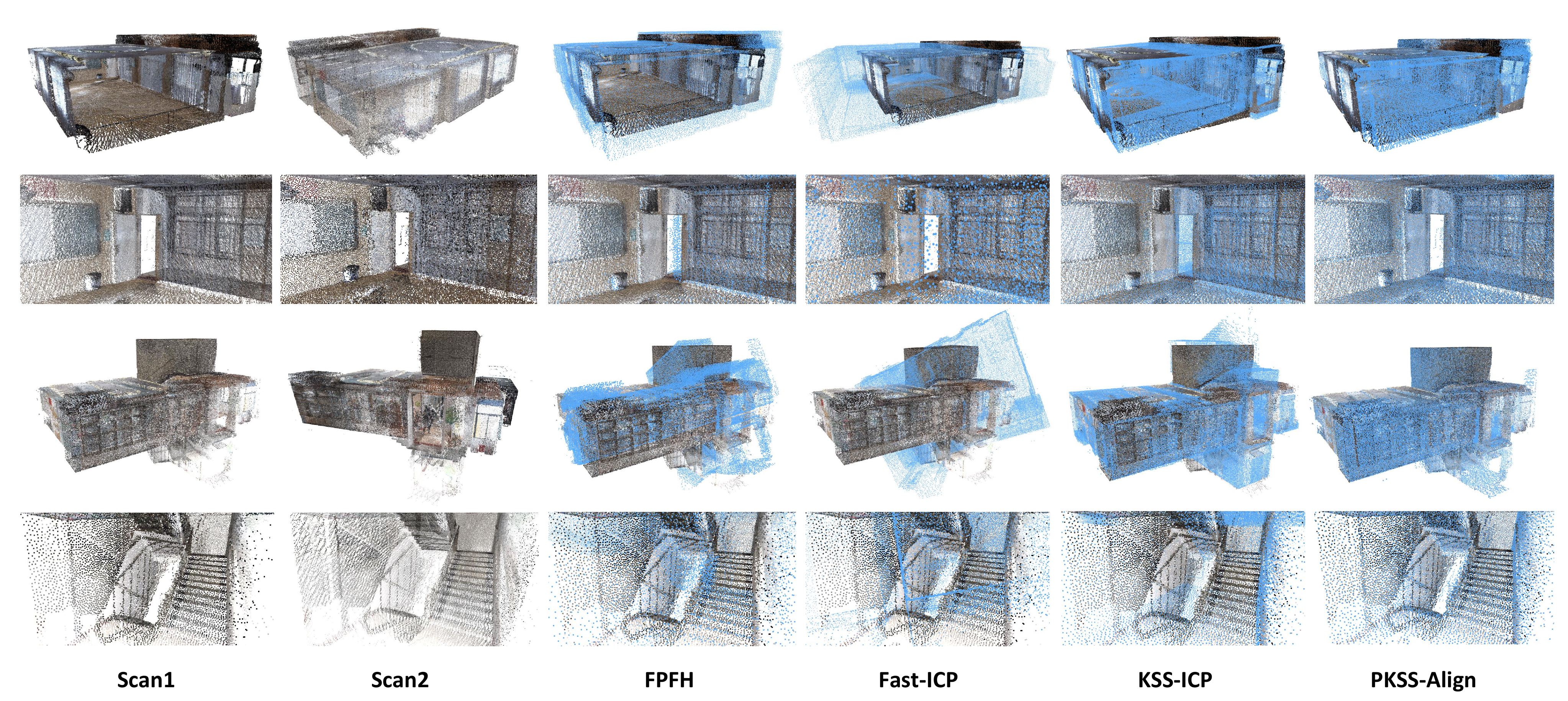}
  \vspace{-3mm}
  \caption{Registration results in the real scene. The registration methods align scan2 to scan1. Point clouds are scanned by LiDAR device.}
  \vspace{-3mm}
  \label{f8_1}
\end{figure*}

\textbf{Robustness to defective parts and non-uniform densities.} As shown in Fig.~\ref{f5_1}, we generate ModelNet40 and S3DIS subsets with similarity transformations, defective parts, and non-uniform densities to evaluate the robustness of our method. Firstly, we artificially delete some regions of point clouds to generate defective parts that are controlled between 30\% and 50\% of global ones. For non-uniform densities generation, we compute the bounding box of point cloud and delete points according to a fixed step. Then, we achieve point clouds with band distributions that simulate the interlaced scanning data. Based on the subsets, we evaluate the performance of different registration methods. In Fig.~\ref{f7}, Table~\ref{t2} and Table~\ref{t2_2}, we visualize some registration instances and report related quantitative results. Since the performance of scale normalization depends on the complete geometric structure, the defective parts bring some uncontrollable degradation in registration. Influenced by the condition, the GO-ICP cannot be converged with a reasonable time cost. To extent the searching region on $SO(3)$, the KSS-ICP increases the time cost obviously. Limited by the Hausdorff metric, the performance of KSS-ICP is further reduced for defective point clouds. In comparison, the PKSS-Align overcomes the limitation with PKSS-based shape measurement and optimized parallel acceleration. 

\textbf{Robustness to noisy points.} In real scenes, the raw point clouds take random noisy points produced by particulates and low precision scanning equipment. Therefore, the registration method should robust to noisy points in a certain degree. To evaluate the robustness, we generate some subsets with different kinds of noisy points based on S3DIS dataset, which are used to simulate real random noise. In order to approximate real situations, we applied the same defect operation as mentioned in previous subsection to all noisy point clouds, which is to simulate obstruction. According to the different distributions, we add two kinds of noise into point clouds, including Gaussian and mean noise. The Gaussian noise is generated by normal distribution $N(0, \sigma)$. The generation is formulated as
\begin{equation}
\label{e_noise}
\begin{array}{c}
p_i'=p_i+n_i\cdot m_i,\;\\
m_i\in\{m\},\{m\}\sim N(0,\sigma^2),\sigma=r\ast l_k,
\end{array}
\end{equation}
where $p_i’$ is the noisy point that is computed from the original point $p_i$ with a random movement $m_i$ according to the normal vector $n_i$. The values of $\{m\}$ satisfy the normal distribution. The $\sigma$ is the distributed control parameter that is computed by the input noisy range $r$ and $l_k$ ($l_k$ is the average length between points and their $k$ neighbors, $k = 12$ by default). The range $r$ controls the value of $\sigma$ that reflects the degree of the noise. We generate three kinds of Gaussian noise with different ranges (0.2, 0.4 and 0.6). Replacing the distribution to uniform one $U(0, \sigma)$, we get test datasets with mean noise. We also use three values of $r$ (0.2, 0,4 and 0.6) to construct three kinds of mean noise. In Fig.~\ref{f8} and Table~\ref{t3}, we show some registration results and report quantitative data for the methods in noisy point clouds. The Go-ICP and Robust-ICP cannot achieve convergence results with an acceptable time cost ($<5min$). For Geo-Transformer, the performance is limited by complex parameters and pre-training weights. In comparison, our method is able to achieve better results. 

\textcolor{black}{It should be noted that some methods show lower performance than the reported in original papers. The reason is that we introduced scale variations in the dataset. Even with PCA-based scale normalization, the performance of general methods tends to degrade under the combined influence of large pose differences. Some methods achieve relatively better MSE values but perform poorly in RR metrics (Go-ICP in Table~\ref{t1}). It means that target point clouds contain symmetrical structures, causing optimization based on point-to-point distances to fail in achieving correct pose alignment. This indirectly demonstrates the effectiveness of shape measurement in registration tasks.}

\textbf{Evaluation on real scanning data.} The mentioned point clouds for registration are generated with artificially interference factors like Gaussian noisy points and specific defective parts. In real scenarios, the raw scanning data contains more random points ans sub-structures. Therefore, the registration performance should be evaluated on real scanning data. We report the measurement results based on 3DMatch and KITTI in Fig~\ref{f8_0_0} and Table~\ref{t2_3}. They take point clouds with significant defective parts and sparse densities. We also collect a small dataset with 46 scanning point clouds that achieved by handheld laser scanner, mobile phone with structured light scanning camera, and drone. Some instances are shown in Fig.\ref{f8_0} and Fig.\ref{f8_1}. The quantitative results are reported in Table~\ref{t4_1}. For real scanning data, most methods cannot achieve accurate results with reasonable convergence speed and computational overhead. Few methods can achieve results efficiently, but the performance is unstable. In contrast, our method achieves better registration accuracy with higher efficiency.

\subsection{Analysis}\label{Quantitative Analysis}

\textbf{Complexity.} We have reported the time cost of different methods in Tables~\ref{t1}-\ref{t4}. The deep encoding-based methods achieve improvement for registration with $O(N)$ complexity, including the encoding of the point clouds and the pose estimation are approximately linear calculations. However, the accuracy of registration is not stable. Such methods are sensitive to the large rotations. RoITr enhances performance across various poses. However, it still struggles with non-uniform scales. The complexity of ICP is $O(N^2)$ that is related to the input point number and initial poses. The Go-ICP is a global searching strategy that is similar to our method. The drawback is the lower computational efficiency. In addition, the convergence cannot be guaranteed when dealing with point clouds with obvious scale differences and large volume. The FR-ICP improves the robustness of original ICP for point clouds with complex similarity transformations, but the performances are sensitive to the defective parts and noisy points. As the state of the art, the MaxCliq achieves fast computation speed and better accuracy for point clouds with lower overlap regions. However, the used graph searching strategy may fail to converge with a certain probability. In practice, the registration process is crash for some point clouds. The limitations of KSS-ICP have been discussed before. To handle the limitations, the PKSS-Align redesign the shape measurement and global searching strategy which improve the accuracy for scale estimation, shape alignment, and computational efficiency. Especially for point clouds with different scales and defective parts at the same time, PKSS-Align achieves more stable performance to support application in actual scenarios.

\begin{figure}
  \centering
  \includegraphics[width=\linewidth]{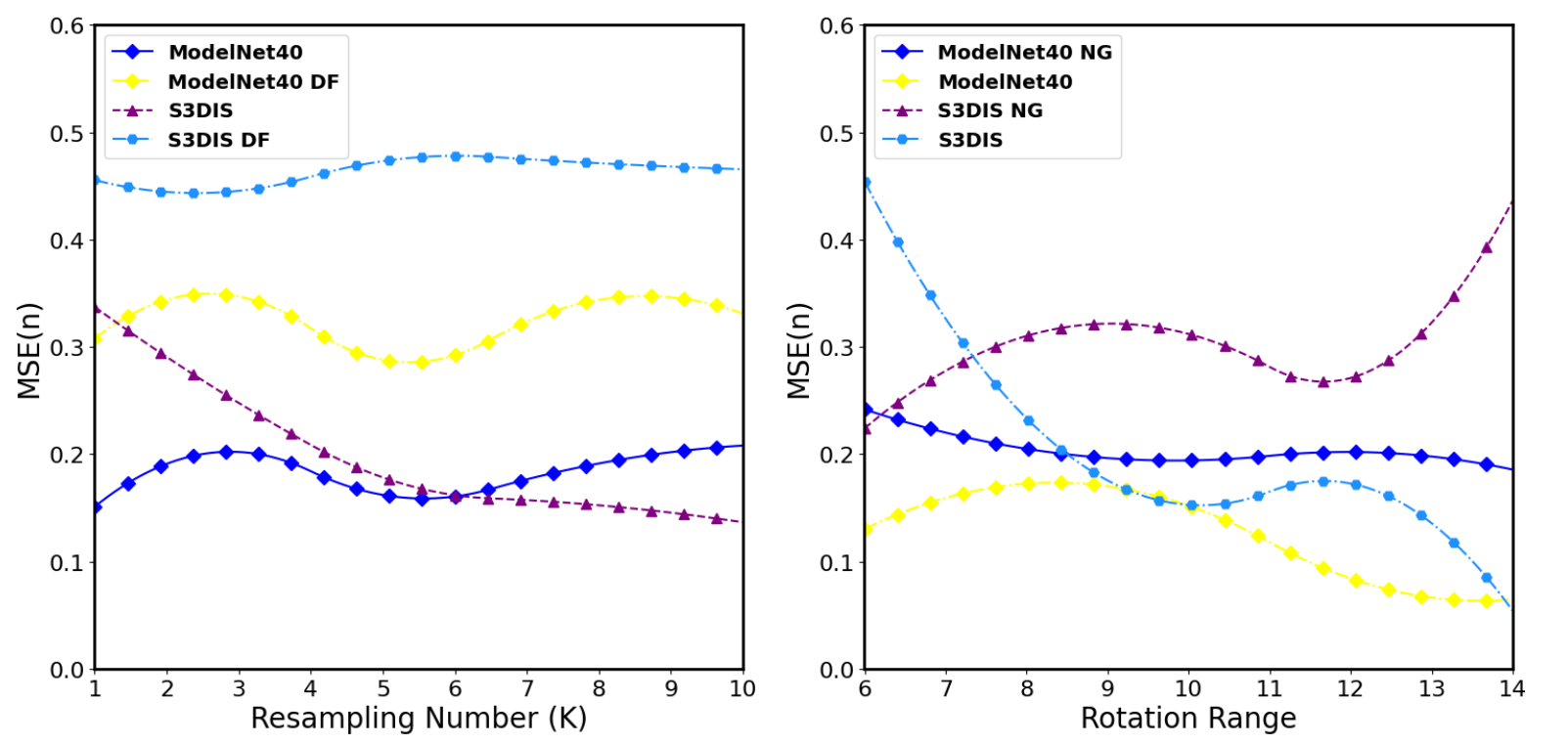}
  \vspace{-3mm}
  \caption{MSE(n)-based curves generated by related parameters. Left: PKSS-Align registration results with related resampling numbers in different test datasets ("DF" means dataset with defective parts); right: PKSS-Align registration results with related rotation ranges ($\sqrt[3]{\vert O_c\vert}$) in different test datasets ("NG" means the geometric feature points are not considered).}
  \vspace{-3mm}
  \label{f9}
\end{figure}

\begin{table}[]
\centering
\caption{Evaluation in real scanning dataset.} 
\label{t4_1}
\begin{tabular}{l|ccc}
\toprule
\textbf{Dataset}                      & \multicolumn{3}{c}{\textbf{Real Scanning Dataset}}    \\ \midrule
\textbf{Method\textbackslash{}Metric} & \textbf{Time$\downarrow$}   & \textbf{MSE$\downarrow$}      & \textbf{MSE(n)$\downarrow$} \\ \midrule
\textbf{FPFH~\cite{yang2016fast}}                         & 39.433s         & 3.41E+05          & 0.7671          \\
\textbf{Fast-ICP~\cite{Zhang2022FastICP}}                     & 17.025s         & 3.90E+05          & 0.7412          \\
\textbf{KSS-ICP~\cite{lv2023kss}}                      & 7.314s          & 3.07E+03          & 0.4422          \\
\textbf{PKSS-Align}                   & 	\cellcolor{lightgray}{\textbf{7.289s}} & \cellcolor{lightgray}{\textbf{1.56E+02}} & \cellcolor{lightgray}{\textbf{0.2856}}\\ \bottomrule 
\end{tabular}
\vspace{-3mm}
\end{table}

\begin{table}[]
\centering
\caption{Evaluations of PKSS-Align on S3DIS with different parameter sets. Translation $\vert S_c\vert$ is the number $\{S_c\}$ mentioned in Eq.~\eqref{e11}, Rotation $\vert O_c\vert$ is the number $\{O_c\}$ mentioned in Eq.~\eqref{e10}, Resampling number means the value of $m$ by resampling method mentioned in Eq.~\eqref{e2}.} 
\vspace{-3mm}
\label{t5}
\resizebox{\linewidth}{!}{
\begin{tabular}{c|c|c||ccc}
\toprule
\textbf{Translation}                 & \textbf{Rotation}                & \textbf{Resampling} & \textbf{Time} & \textbf{MSE} & \textbf{MSE(n)} \\ \midrule
\multirow{9}{*}{\textbf{$\vert S_c\vert = 3^3$}} & \multirow{3}{*}{\textbf{$\vert O_c\vert = 10^3$}} & \textbf{1,000}     & 9.874s        & 0.0512       & 0.5071          \\
                                &                                  & \textbf{3,000}     & 13.765s       & 0.0653       & 0.4754          \\
                                &                                  & \textbf{5,000}     & 16.644s       & 0.0576       & 0.4425          \\
                                & \multirow{3}{*}{\textbf{$\vert  O_c\vert = 12^3$}} & \textbf{1,000}     & 7.562s        & 0.0485       & 0.4555          \\
                                &                                  & \textbf{3,000}     & 11.892s       & 0.0451       & 0.4454          \\
                                &                                  & \textbf{5,000}     & 27.122s       & 0.0433       & 0.4731          \\
                                & \multirow{3}{*}{\textbf{$\vert  O_c\vert = 14^3$}} & \textbf{1,000}     & 11.096s       & 0.0492       & 0.3898          \\
                                &                                  & \textbf{3,000}     & 11.765s       & 0.0607       & 0.3915          \\
                                &                                  & \textbf{5,000}     & 17.895s       & 0.0501       & 0.4242          \\ \midrule
\multirow{9}{*}{\textbf{$\vert S_c\vert = 5^3$}} & \multirow{3}{*}{\textbf{$\vert  O_c\vert = 10^3$}} & \textbf{1,000}     & 10.075s       & 0.0494       & 0.5199          \\
                                &                                  & \textbf{3,000}     & 19.029s       & 0.0664       & 0.5136          \\
                                &                                  & \textbf{5,000}     & 20.459s       & 0.0562       & 0.4273          \\
                                & \multirow{3}{*}{\textbf{$\vert  O_c\vert = 12^3$}} & \textbf{1,000}     & 5.562s        & 0.0543       & 0.4627          \\
                                &                                  & \textbf{3,000}     & 11.751s       & 0.0337       & 0.4172          \\
                                &                                  & \textbf{5,000}     & 18.042s       & 0.0385      & 0.4487          \\
                                & \multirow{3}{*}{\textbf{$\vert  O_c\vert = 14^3$}} & \textbf{1,000}     & 5.832s        & 0.0368     & 0.4108          \\
                                &                                  & \textbf{3,000}     & 24.343s       & 0.0456       & 0.3864          \\
                                &                                  & \textbf{5,000}     & 24.417s       & 0.0455       & 0.3769   \\ \bottomrule      
\end{tabular}
}
\vspace{-2mm}
\end{table}

\begin{table}[t]\color{black}
\centering
\caption{Evaluations of PKSS-Align on S3DIS with and without related modules (Measurement: PKSS-based shape measurement, Global: global searching scheme). } 
\label{t9}
\vspace{-2mm}
\begin{tabular}{cc|ccc}
\toprule
\textbf{Measurement} & \textbf{Global} & \textbf{MSE}$\downarrow$ & \textbf{MSE(n)}$\downarrow$ & \textbf{${\boldsymbol {GT}}_{\boldsymbol cos}$}$\uparrow$ \\ \midrule
 \checkmark   &   & 0.07339   & 0.6327     & 0.4852   \\
  &  \checkmark   & 0.00871   & 0.1353    & 0.4498   \\
 \checkmark  &  \checkmark  & 0.00493    & 0.1187   & 0.8826 \\ \bottomrule                                       
\end{tabular}
\vspace{-3mm}
\end{table}

\textbf{Ablation.} Some parameters related to the PKSS-Align should be discussed, including resampling, rotation and translation parameters. The resampling mentioned in pre-processing is to balance the efficiency and precision. The higher resampling rate produces redundant points that reduce the efficiency. On the contrary, insufficient points break the shape feature of original point cloud and reduce the accurate of PKSS-based shape measurement. We compute the MSE-based curves with different resampling numbers. The results are visualized in Fig.~\ref{f9} with quantitative analysis in Table~\ref{t5}. It cannot be regarded as a corresponding simplification of the original point cloud when there are too few sampling points. More shape features are ignored. On the contrary, too many sampling points increase the computational cost without significant improvement for accuracy. 

Based on the experimental results, the proper resampling number should be controlled around $3k\sim 5k$ for the balance. The rotation and translation parameters mentioned in Sec.~\ref{Global Searching Strategy} also take important influence in registration. Similar to the resampling, larger values of the parameters support more accurate global searching based on the PKSS-based shape measurement. \textcolor{black}{Benefited from the GPU-based acceleration, the time cost does not significantly increase with $|O_c|$.} In Fig.~\ref{f9} and Table~\ref{t5}, we compare the MSE-based curves and related metrics with different numerical combinations of the parameters in ModelNet40 and S3DIS test datasets. For point clouds with symmetric structures, more rotation searching cannot improve the accuracy without internal geometric feature analysis. Once internal geometric features are added into the shape measurement, the registration accuracy and rotation searching show a significant proportional relationship. Based on the above analysis, we set the default parameters mentioned in Sec.~\ref{Global Searching Strategy}.

\textcolor{black}{To better illustrate the roles of different components in PKSS-Align, we additionally compare the registration performance under various module combinations. Table~\ref{t9} shows registration results based on S3DIS. Firstly, we report the registration result by PKSS-based shape measurement without global searching scheme. This reduces our method to a local registration scheme, approximating the original ICP. Then, We use point-based metric to instead the PKSS-based shape measurement. It reduces the proposed method to its earlier version (KSS-ICP).}

\begin{figure}
  \centering
  \includegraphics[width=\linewidth]{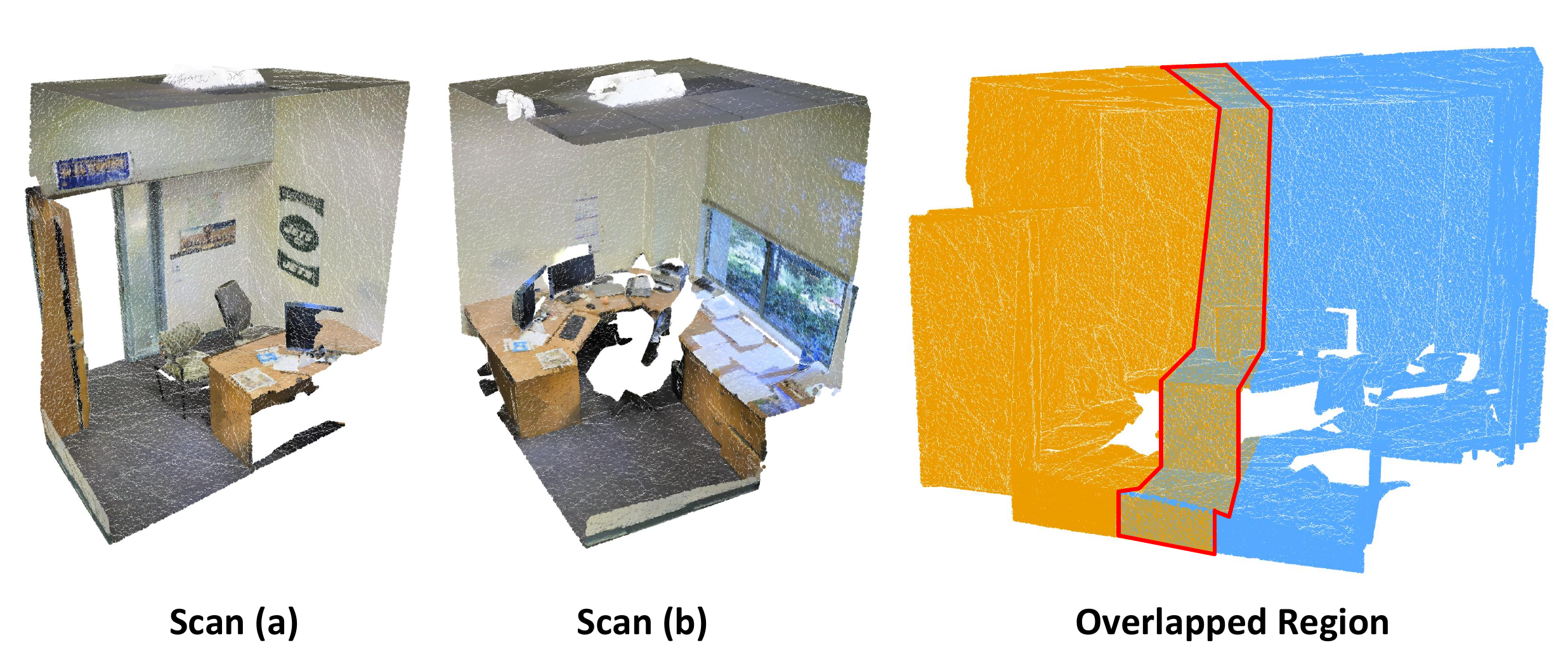}
  \vspace{-3mm}
  \caption{Instance of low overlapped point cloud registration. Scan (a) and scan (b) are two point clouds scanned from same scene, the yellow lines label the overlapped region between (a) and (b).}
  \vspace{-3mm}
  \label{f11}
\end{figure}

\begin{table}[t]\color{black}
\centering
\caption{Evaluations of PKSS-Align on low overlapped point clouds of 3DMatch} 
\label{t10}
\vspace{-2mm}
\begin{tabular}{l|cccc}
\toprule
\textbf{Methods}  & \textbf{MSE}$\downarrow$ & \textbf{MSE(n)}$\downarrow$ &  \textbf{${\boldsymbol {GT}}_{\boldsymbol cos}$}$\uparrow$ & \textbf{RR}$\uparrow$\\ \midrule
GeoT~\cite{qin2022geometric}  & 0.40861	& 0.7162  & 0.3034  & $<$10\%  \\
PKSS-Align & 0.02914  & 0.8706 & 0.5545 & 42\%   \\ 
\bottomrule 
\end{tabular}
\vspace{-3mm}
\end{table}

\textbf{Limitations.} In essence, PKSS-Align is looking for a contour-based shape matching based on the discrete form of point cloud. The contour is constructed by representative samples while considering the internal geometric features. For low overlapped point clouds that are shown in Fig.~\ref{f11}, the performance of PKSS-Align will degrade with high probability. \textcolor{black}{Table~\ref{t10} shows quantitative results for low overlapped point clouds from 3DMatch (overlap region is less than 50\%). Even with the global searching scheme, the registration recall degraded to below 45\%.} The reason is that there is no global correspondence between low overlapped point clouds according to the contour. Forcibly using PKSS-Align for registration will only yield a local matching result. Nevertheless, PKSS-Align is already capable of handling the majority of registration tasks in practice. It can achieve accurate registration results for point clouds with more than half overlapped regions while taking defective parts, larger pose and scale differences, and noisy points.

\section{Conclusions}

In this paper, we propose a robust point cloud registration method PKSS-Align that is implemented on Pre-Kendall shape space. It estimates the transformation matrix between point clouds by PKSS-based shape measurement that is robust to similarity transformations, non-uniform densities, and noisy points. With a practical implementation of global searching strategy, the proposed registration also can align point clouds with defective parts while keeping the efficiency. It does not require additional deep feature-encoding and data training. Experiments show that our method has good performance in different registration tasks. In future work, we consider combining the overlapped region searching in our framework. With the overlapped regions, the PKSS-Align can handle various registration tasks.

\ifCLASSOPTIONcaptionsoff
  \newpage
\fi

\bibliographystyle{IEEEtran}
\bibliography{IEEEabrv,IEEEBib}

\begin{thebibliography}{10}
\providecommand{\url}[1]{#1}
\csname url@samestyle\endcsname
\providecommand{\newblock}{\relax}
\providecommand{\bibinfo}[2]{#2}
\providecommand{\BIBentrySTDinterwordspacing}{\spaceskip=0pt\relax}
\providecommand{\BIBentryALTinterwordstretchfactor}{4}
\providecommand{\BIBentryALTinterwordspacing}{\spaceskip=\fontdimen2\font plus
\BIBentryALTinterwordstretchfactor\fontdimen3\font minus
  \fontdimen4\font\relax}
\providecommand{\BIBforeignlanguage}[2]{{%
\expandafter\ifx\csname l@#1\endcsname\relax
\typeout{** WARNING: IEEEtran.bst: No hyphenation pattern has been}%
\typeout{** loaded for the language `#1'. Using the pattern for}%
\typeout{** the default language instead.}%
\else
\language=\csname l@#1\endcsname
\fi
#2}}
\providecommand{\BIBdecl}{\relax}
\BIBdecl

\bibitem{besl1992method}
P.~J. Besl and N.~D. McKay, ``Method for registration of 3-d shapes,'' in
  \emph{Sensor fusion IV: control paradigms and data structures}, vol.
  1611.\hskip 1em plus 0.5em minus 0.4em\relax International Society for Optics
  and Photonics, 1992, pp. 586--606.

\bibitem{park2003accurate}
S.-Y. Park and M.~Subbarao, ``An accurate and fast point-to-plane registration
  technique,'' \emph{Pattern Recognition Letters}, vol.~24, no.~16, pp.
  2967--2976, 2003.

\bibitem{wahba1965least}
G.~Wahba, ``A least squares estimate of satellite attitude,'' \emph{SIAM
  review}, vol.~7, no.~3, pp. 409--409, 1965.

\bibitem{rusu2009fast}
R.~B. Rusu, N.~Blodow, and M.~Beetz, ``Fast point feature histograms (fpfh) for
  3d registration,'' in \emph{IEEE Int. Conf. on robotics and
  automation}.\hskip 1em plus 0.5em minus 0.4em\relax IEEE, 2009, pp.
  3212--3217.

\bibitem{horn1988closed}
B.~K. Horn, H.~M. Hilden, and S.~Negahdaripour, ``Closed-form solution of
  absolute orientation using orthonormal matrices,'' \emph{JOSA A}, vol.~5,
  no.~7, pp. 1127--1135, 1988.

\bibitem{aoki2019pointnetlk}
Y.~Aoki, H.~Goforth, R.~A. Srivatsan, and S.~Lucey, ``Pointnetlk: Robust \&
  efficient point cloud registration using pointnet,'' in \emph{Proc. IEEE/CVF
  Conf. on Computer Vision \& Pattern Recognition}, 2019, pp. 7163--7172.

\bibitem{sarode2019pcrnet}
V.~Sarode, X.~Li, H.~Goforth, Y.~Aoki, R.~A. Srivatsan, S.~Lucey, and
  H.~Choset, ``Pcrnet: Point cloud registration network using pointnet
  encoding,'' \emph{arXiv preprint arXiv:1908.07906}, 2019.

\bibitem{lv2023kss}
C.~Lv, W.~Lin, and B.~Zhao, ``Kss-icp: Point cloud registration based on
  kendall shape space,'' \emph{IEEE Trans. on Image Processing}, vol.~32, pp.
  1681--1693, 2023.

\bibitem{kendall1984shape}
D.~G. Kendall, ``Shape manifolds, procrustean metrics, and complex projective
  spaces,'' \emph{Bulletin of the London mathematical society}, vol.~16, no.~2,
  pp. 81--121, 1984.

\bibitem{fitzgibbon2003robust}
A.~W. Fitzgibbon, ``Robust registration of 2d and 3d point sets,'' \emph{Image
  and Vision Computing}, vol.~21, no. 13-14, pp. 1145--1153, Dec. 2003.

\bibitem{granger2002multi}
S.~Granger and X.~Pennec, ``Multi-scale em-icp: A fast and robust approach for
  surface registration,'' in \emph{Proc. Euro. Conf. on Computer Vision}.\hskip
  1em plus 0.5em minus 0.4em\relax Springer, 2002, pp. 418--432.

\bibitem{ying2009scale}
S.~Ying, J.~Peng, S.~Du, and H.~Qiao, ``A scale stretch method based on icp for
  3d data registration,'' \emph{IEEE Trans. on Automation Science and
  Engineering}, vol.~6, no.~3, pp. 559--565, May. 2009.

\bibitem{jian2005robust}
B.~Jian and B.~C. Vemuri, ``A robust algorithm for point set registration using
  mixture of gaussians,'' in \emph{Proc. Int. Conf. on Computer Vision},
  vol.~2.\hskip 1em plus 0.5em minus 0.4em\relax IEEE, 2005, pp. 1246--1251.

\bibitem{yang2019point}
Y.~Yang, D.~Fan, S.~Du, M.~Wang, B.~Chen, and Y.~Gao, ``Point set registration
  with similarity and affine transformations based on bidirectional kmpe
  loss,'' \emph{IEEE trans. on Cybernetics}, vol.~51, no.~3, pp. 1678--1689,
  Oct. 2019.

\bibitem{Zhang2022FastICP}
J.~Zhang, Y.~Yao, and B.~Deng, ``Fast and robust iterative closest point,''
  \emph{IEEE Trans. Pattern Analysis \& Machine Intelligence}, vol.~44, no.~7,
  pp. 3450--3466, Jul. 2022.

\bibitem{makela2002review}
T.~Makela, P.~Clarysse, O.~Sipila, N.~Pauna, Q.~C. Pham, T.~Katila, and I.~E.
  Magnin, ``A review of cardiac image registration methods,'' \emph{IEEE Trans.
  on Medical Imaging}, vol.~21, no.~9, pp. 1011--1021, Sep. 2002.

\bibitem{rusinkiewicz2001efficient}
S.~Rusinkiewicz and M.~Levoy, ``Efficient variants of the icp algorithm,'' in
  \emph{Proc. Int. Conf. on 3-D Digital Imaging and Modeling}.\hskip 1em plus
  0.5em minus 0.4em\relax IEEE, 2001, pp. 145--152.

\bibitem{olsson2008branch}
C.~Olsson, F.~Kahl, and M.~Oskarsson, ``Branch-and-bound methods for euclidean
  registration problems,'' \emph{IEEE Trans. Pattern Analysis \& Machine
  Intelligence}, vol.~31, no.~5, pp. 783--794, May. 2008.

\bibitem{li20073d}
H.~Li and R.~Hartley, ``The 3d-3d registration problem revisited,'' in
  \emph{Proc. Int. Conf. on Computer Vision}.\hskip 1em plus 0.5em minus
  0.4em\relax IEEE, 2007, pp. 1--8.

\bibitem{parra2014fast}
A.~Parra~Bustos, T.-J. Chin, and D.~Suter, ``Fast rotation search with
  stereographic projections for 3d registration,'' in \emph{Proc. IEEE/CVF
  Conf. on Computer Vision \& Pattern Recognition}, 2014, pp. 3930--3937.

\bibitem{bazin2012globally}
J.-C. Bazin, Y.~Seo, and M.~Pollefeys, ``Globally optimal consensus set
  maximization through rotation search,'' in \emph{Asian Conf. on Computer
  Vision}.\hskip 1em plus 0.5em minus 0.4em\relax Springer, 2012, pp. 539--551.

\bibitem{enqvist2008robust}
O.~Enqvist and F.~Kahl, ``Robust optimal pose estimation,'' in \emph{Proc.
  Euro. Conf. on Computer Vision}.\hskip 1em plus 0.5em minus 0.4em\relax
  Springer, 2008, pp. 141--153.

\bibitem{yang2013go}
J.~Yang, H.~Li, and Y.~Jia, ``Go-icp: Solving 3d registration efficiently and
  globally optimally,'' in \emph{Proc. Int. Conf. on Computer Vision}, 2013,
  pp. 1457--1464.

\bibitem{biber2003normal}
P.~Biber and W.~Stra{\ss}er, ``The normal distributions transform: A new
  approach to laser scan matching,'' in \emph{IEEE/RSJ Int. Conf. on
  Intelligent Robots and Systems}, vol.~3.\hskip 1em plus 0.5em minus
  0.4em\relax IEEE, 2003, pp. 2743--2748.

\bibitem{serafin2015nicp}
J.~Serafin and G.~Grisetti, ``Nicp: Dense normal based point cloud
  registration,'' in \emph{IEEE/RSJ Int. Conf. on Intelligent Robots and
  Systems}.\hskip 1em plus 0.5em minus 0.4em\relax IEEE, 2015, pp. 742--749.

\bibitem{belongie2002shape}
S.~Belongie, J.~Malik, and J.~Puzicha, ``Shape matching and object recognition
  using shape contexts,'' \emph{IEEE Trans. Pattern Analysis \& Machine
  Intelligence}, vol.~24, no.~4, pp. 509--522, Apr. 2002.

\bibitem{frome2004recognizing}
A.~Frome, D.~Huber, R.~Kolluri, T.~B{\"u}low, and J.~Malik, ``Recognizing
  objects in range data using regional point descriptors,'' in \emph{Proc.
  Euro. Conf. on Computer Vision}.\hskip 1em plus 0.5em minus 0.4em\relax
  Springer, 2004, pp. 224--237.

\bibitem{guan2009registration}
T.~Guan and C.~Wang, ``Registration based on scene recognition and natural
  features tracking techniques for wide-area augmented reality systems,''
  \emph{IEEE Trans. on Multimedia}, vol.~11, no.~8, pp. 1393--1406, Sep. 2009.

\bibitem{guo2014accurate}
Y.~Guo, F.~Sohel, M.~Bennamoun, J.~Wan, and M.~Lu, ``An accurate and robust
  range image registration algorithm for 3d object modeling,'' \emph{IEEE
  Trans. on Multimedia}, vol.~16, no.~5, pp. 1377--1390, Apr. 2014.

\bibitem{tabia2015covariance}
H.~Tabia and H.~Laga, ``Covariance-based descriptors for efficient 3d shape
  matching, retrieval, and classification,'' \emph{IEEE Trans. on Multimedia},
  vol.~17, no.~9, pp. 1591--1603, Jul. 2015.

\bibitem{rusu2008aligning}
R.~B. Rusu, N.~Blodow, Z.~C. Marton, and M.~Beetz, ``Aligning point cloud views
  using persistent feature histograms,'' in \emph{IEEE/RSJ Int. Conf. on
  intelligent robots and systems}.\hskip 1em plus 0.5em minus 0.4em\relax IEEE,
  2008, pp. 3384--3391.

\bibitem{yang2016fast}
J.~Yang, Z.~Cao, and Q.~Zhang, ``A fast and robust local descriptor for 3d
  point cloud registration,'' \emph{Information Sciences}, vol. 346, pp.
  163--179, Mar. 2016.

\bibitem{chen2022sc2}
Z.~Chen, K.~Sun, F.~Yang, and W.~Tao, ``Sc2-pcr: A second order spatial
  compatibility for efficient and robust point cloud registration,'' in
  \emph{Proc. IEEE/CVF Conf. on Computer Vision \& Pattern Recognition}, 2022,
  pp. 13\,221--13\,231.

\bibitem{zhang20233d}
X.~Zhang, J.~Yang, S.~Zhang, and Y.~Zhang, ``3d registration with maximal
  cliques,'' in \emph{Proc. IEEE/CVF Conf. on Computer Vision \& Pattern
  Recognition}, 2023, pp. 17\,745--17\,754.

\bibitem{yuan2024inlier}
Y.~Yuan, Y.~Wu, X.~Fan, M.~Gong, Q.~Miao, and W.~Ma, ``Inlier confidence
  calibration for point cloud registration,'' in \emph{Proc. IEEE/CVF Conf. on
  Computer Vision \& Pattern Recognition}, 2024, pp. 5312--5321.

\bibitem{liu2024extend}
Q.~Liu, H.~Zhu, Z.~Wang, Y.~Zhou, S.~Chang, and M.~Guo, ``Extend your own
  correspondences: Unsupervised distant point cloud registration by progressive
  distance extension,'' in \emph{Proc. IEEE/CVF Conf. on Computer Vision \&
  Pattern Recognition}, 2024, pp. 20\,816--20\,826.

\bibitem{qi2017pointnet}
C.~R. Qi, H.~Su, K.~Mo, and L.~J. Guibas, ``Pointnet: Deep learning on point
  sets for 3d classification and segmentation,'' in \emph{Proc. IEEE/CVF Conf.
  on Computer Vision \& Pattern Recognition}, 2017, pp. 652--660.

\bibitem{qi2017pointnet++}
C.~R. Qi, L.~Yi, H.~Su, and L.~J. Guibas, ``Pointnet++ deep hierarchical
  feature learning on point sets in a metric space,'' in \emph{Proc. Conf. on
  Neural Information Processing Systems}, 2017, pp. 5105--5114.

\bibitem{lu2019deepicp}
W.~Lu, G.~Wan, Y.~Zhou, X.~Fu, P.~Yuan, and S.~Song, ``Deepicp: An end-to-end
  deep neural network for 3d point cloud registration,'' \emph{arXiv preprint
  arXiv:1905.04153}, 2019.

\bibitem{wang2019deep}
Y.~Wang and J.~M. Solomon, ``Deep closest point: Learning representations for
  point cloud registration,'' in \emph{Proc. IEEE/CVF Conf. on Computer Vision
  \& Pattern Recognition}, 2019, pp. 3523--3532.

\bibitem{wang2019prnet}
------, ``Prnet: Self-supervised learning for partial-to-partial
  registration,'' \emph{arXiv preprint arXiv:1910.12240}, 2019.

\bibitem{li2019iterative}
J.~Li, C.~Zhang, Z.~Xu, H.~Zhou, and C.~Zhang, ``Iterative distance-aware
  similarity matrix convolution with mutual-supervised point elimination for
  efficient point cloud registration,'' \emph{arXiv preprint arXiv:1910.10328},
  2019.

\bibitem{yew2020rpm}
Z.~J. Yew and G.~H. Lee, ``Rpm-net: Robust point matching using learned
  features,'' in \emph{Proc. IEEE/CVF Conf. on Computer Vision \& Pattern
  Recognition}, 2020, pp. 11\,824--11\,833.

\bibitem{pais20203dregnet}
G.~D. Pais, S.~Ramalingam, V.~M. Govindu, J.~C. Nascimento, R.~Chellappa, and
  P.~Miraldo, ``3dregnet: A deep neural network for 3d point registration,'' in
  \emph{Proc. IEEE/CVF Conf. on Computer Vision \& Pattern Recognition}, 2020,
  pp. 7193--7203.

\bibitem{choy2020deep}
C.~Choy, W.~Dong, and V.~Koltun, ``Deep global registration,'' in \emph{Proc.
  IEEE/CVF Conf. on Computer Vision \& Pattern Recognition}, 2020, pp.
  2514--2523.

\bibitem{gu2022rcp}
X.~Gu, C.~Tang, W.~Yuan, Z.~Dai, S.~Zhu, and P.~Tan, ``Rcp: Recurrent closest
  point for point cloud,'' in \emph{Proc. IEEE/CVF Conf. on Computer Vision \&
  Pattern Recognition}, 2022, pp. 8216--8226.

\bibitem{qin2022geometric}
Z.~Qin, H.~Yu, C.~Wang, Y.~Guo, Y.~Peng, and K.~Xu, ``Geometric transformer for
  fast and robust point cloud registration,'' in \emph{Proc. IEEE/CVF Conf. on
  Computer Vision \& Pattern Recognition}, 2022, pp. 11\,143--11\,152.

\bibitem{wang2023roreg}
H.~Wang, Y.~Liu, Q.~Hu, B.~Wang, J.~Chen, Z.~Dong, Y.~Guo, W.~Wang, and
  B.~Yang, ``Roreg: Pairwise point cloud registration with oriented descriptors
  and local rotations,'' \emph{IEEE Trans. Pattern Analysis \& Machine
  Intelligence}, vol.~45, no.~8, pp. 10\,376--10\,393, 2023.

\bibitem{mei2021point}
G.~Mei, ``Point cloud registration with self-supervised feature learning and
  beam search,'' in \emph{Digital Image Computing: Techniques and
  Applications}.\hskip 1em plus 0.5em minus 0.4em\relax IEEE, 2021, pp. 01--08.

\bibitem{fu2021robust}
K.~Fu, S.~Liu, X.~Luo, and M.~Wang, ``Robust point cloud registration framework
  based on deep graph matching,'' in \emph{Proc. IEEE/CVF Conf. on Computer
  Vision \& Pattern Recognition}, 2021, pp. 8893--8902.

\bibitem{yew2022regtr}
Z.~J. Yew and G.~H. Lee, ``Regtr: End-to-end point cloud correspondences with
  transformers,'' in \emph{Proc. IEEE/CVF Conf. on Computer Vision \& Pattern
  Recognition}, 2022, pp. 6677--6686.

\bibitem{mei2023unsupervised}
G.~Mei, H.~Tang, X.~Huang, W.~Wang, J.~Liu, J.~Zhang, L.~Van~Gool, and Q.~Wu,
  ``Unsupervised deep probabilistic approach for partial point cloud
  registration,'' in \emph{Proc. IEEE/CVF Conf. on Computer Vision \& Pattern
  Recognition}, 2023, pp. 13\,611--13\,620.

\bibitem{yu2023rotation}
H.~Yu, Z.~Qin, J.~Hou, M.~Saleh, D.~Li, B.~Busam, and S.~Ilic,
  ``Rotation-invariant transformer for point cloud matching,'' in \emph{Proc.
  IEEE/CVF Conf. on Computer Vision \& Pattern Recognition}, 2023, pp.
  5384--5393.

\bibitem{jin2024multiway}
S.~Jin, I.~Armeni, M.~Pollefeys, and D.~Barath, ``Multiway point cloud
  mosaicking with diffusion and global optimization,'' in \emph{Proc. IEEE/CVF
  Conf. on Computer Vision \& Pattern Recognition}, 2024, pp. 20\,838--20\,849.

\bibitem{nava2020geodesic}
E.~Nava-Yazdani, H.-C. Hege, T.~J. Sullivan, and C.~von Tycowicz, ``Geodesic
  analysis in kendall’s shape space with epidemiological applications,''
  \emph{Journal of Mathematical Imaging and Vision}, vol.~62, no.~4, pp.
  549--559, 2020.

\bibitem{lv2021approximate}
C.~Lv, W.~Lin, and B.~Zhao, ``Approximate intrinsic voxel structure for point
  cloud simplification,'' \emph{IEEE Trans. on Image Processing}, vol.~30, pp.
  7241--7255, Aug. 2021.

\bibitem{dimitrov2009bounds}
D.~Dimitrov, C.~Knauer, K.~Kriegel, and G.~Rote, ``Bounds on the quality of the
  pca bounding boxes,'' \emph{Computational Geometry}, vol.~42, no.~8, pp.
  772--789, 2009.

\bibitem{wu20153d}
Z.~Wu, S.~Song, A.~Khosla, F.~Yu, L.~Zhang, X.~Tang, and J.~Xiao, ``3d
  shapenets: A deep representation for volumetric shapes,'' in \emph{Proc.
  IEEE/CVF Conf. on Computer Vision \& Pattern Recognition}, 2015, pp.
  1912--1920.

\bibitem{armeni_cvpr16}
I.~Armeni, O.~Sener, A.~R. Zamir, H.~Jiang, I.~Brilakis, M.~Fischer, and
  S.~Savarese, ``3d semantic parsing of large-scale indoor spaces,'' in
  \emph{Proc. IEEE/CVF Conf. on Computer Vision \& Pattern Recognition}, 2016.

\bibitem{zeng20173dmatch}
A.~Zeng, S.~Song, M.~Nie{\ss}ner, M.~Fisher, J.~Xiao, and T.~Funkhouser,
  ``3dmatch: Learning local geometric descriptors from rgb-d reconstructions,''
  in \emph{Proc. IEEE/CVF Conf. on Computer Vision \& Pattern Recognition},
  2017, pp. 1802--1811.

\bibitem{behley2019semantickitti}
J.~Behley, M.~Garbade, A.~Milioto, J.~Quenzel, S.~Behnke, C.~Stachniss, and
  J.~Gall, ``Semantickitti: A dataset for semantic scene understanding of lidar
  sequences,'' in \emph{Proc. IEEE/CVF Conf. on Computer Vision \& Pattern
  Recognition}, 2019, pp. 9297--9307.

\end{thebibliography}

\end{document}